\documentclass[10pt,final,journal]{templates/IEEEtran}

\usepackage[utf8x]{inputenc}
\usepackage[T1]{fontenc}
\usepackage[english]{babel}
\usepackage{amsmath}
\usepackage{amsfonts}
\usepackage{amssymb}
\usepackage{amsthm}
\usepackage{bm}
\usepackage{xspace}
\usepackage{subcaption}
\usepackage[boxruled]{algorithm2e}
\usepackage{mathrsfs}
\usepackage{multirow}
\usepackage{eurosym}
\usepackage{dsfont}
\usepackage[normalem]{ulem} 
\usepackage{balance}
\usepackage{soul}
\usepackage{pdfpages}
\usepackage{graphicx}
\usepackage{todonotes}
\usepackage{url}
\usepackage{units}
\usepackage{siunitx}
\usepackage{pgfplots}
\usepackage[hidelinks]{hyperref}
\usepackage{stfloats}
\usepackage{array}
\usepackage{adjustbox}
\usepackage{float}
\usepackage{enumitem}
\usepackage[normalem]{ulem}
\setitemize{leftmargin=*}
\usepackage{balance}
\newcolumntype{L}[1]{>{\raggedright\let\newline\\\arraybackslash\hspace{0pt}}m{#1}}
\newcolumntype{C}[1]{>{\centering\let\newline\\\arraybackslash\hspace{0pt}}m{#1}}
\newcolumntype{R}[1]{>{\raggedleft\let\newline\\\arraybackslash\hspace{0pt}}m{#1}}

\newcommand{\T}{\mathsf{T}}
\newcommand{\norm}[1]{\left\lVert#1\right\rVert}
\newcommand{\R}{\mathbb{R}}
\newcommand{\N}{\mathbb{N}}
\newcommand{\SE}{SE(3)}
\newcommand{\mat}[1]{\begin{bmatrix}#1\end{bmatrix}}
\newcommand{\bmd}[1]{\bm{\dot{#1}}}
\newcommand{\bmdd}[1]{\bm{\ddot{#1}}}
\newcommand{\fracdiff}[2]{\frac{\partial #1}{\partial #2}}

\definecolor{lightyellow}{rgb}{1.0,0.98,0.7}
\sethlcolor{lightyellow}

\title{Multi-Contact Motion Retargeting using Whole-body Optimization of Full Kinematics and Sequential Force Equilibrium}
\author{Quentin Rouxel, Kai Yuan, Ruoshi Wen, Zhibin Li %Ruoshi Wen,

}

\begin{document}

\maketitle

%%%
%%% Abstract and keywords
%%%

\begin{abstract}
This paper presents a multi-contact motion adaptation framework that enables teleoperation of high degree-of-freedom (DoF) robots, such as quadrupeds and humanoids, for loco-manipulation tasks in multi-contact settings. Our proposed algorithms optimize whole-body configurations and formulate the retargeting of multi-contact motions as sequential quadratic programming, which is robust and stable near the edges of feasibility constraints. Our framework allows real-time operation of the robot and reduces cognitive load for the operator because infeasible commands are automatically adapted into physically stable and viable motions on the robot. The results in simulations with full dynamics demonstrated the effectiveness of teleoperating different legged robots interactively and generating rich multi-contact movements. We evaluated the computational efficiency of the proposed algorithms, and further validated and analyzed multi-contact loco-manipulation tasks on humanoid and quadruped robots by reaching, active pushing and various traversal on uneven terrains. 
\end{abstract}

%%%
%%% Document content
%%%

\section{Introduction}\label{sec:introduction}

Human-in-the-loop approaches for controlling robots are of essential importance in safety-critical, cognitively challenging and high-risk tasks \cite{dedonato2015human, johnson2015team}, which can be achieved through high-level supervision and/or online commands. Human's involvements in the control loop complement robots' abilities in perception and motor actions. It also provides robots with versatile motor skills for unforeseen situations and dexterous interactions in uncertain environments. Contextual understanding and safe decisions are required to deploy robots in remote tasks, such as distant planetary exploration \cite{schmaus2018preliminary}, subsea inspection and nuclear decommissioning. For example, the ESA's METERON project developed a teleoperation system where the operators' skills in dealing with different situations were used to control planetary robots from the orbit \cite{de2011set}.

\begin{figure}[t]
    \centering
    \begin{subfigure}{0.4\linewidth}
		\centering
		\includegraphics[height=3.5cm]{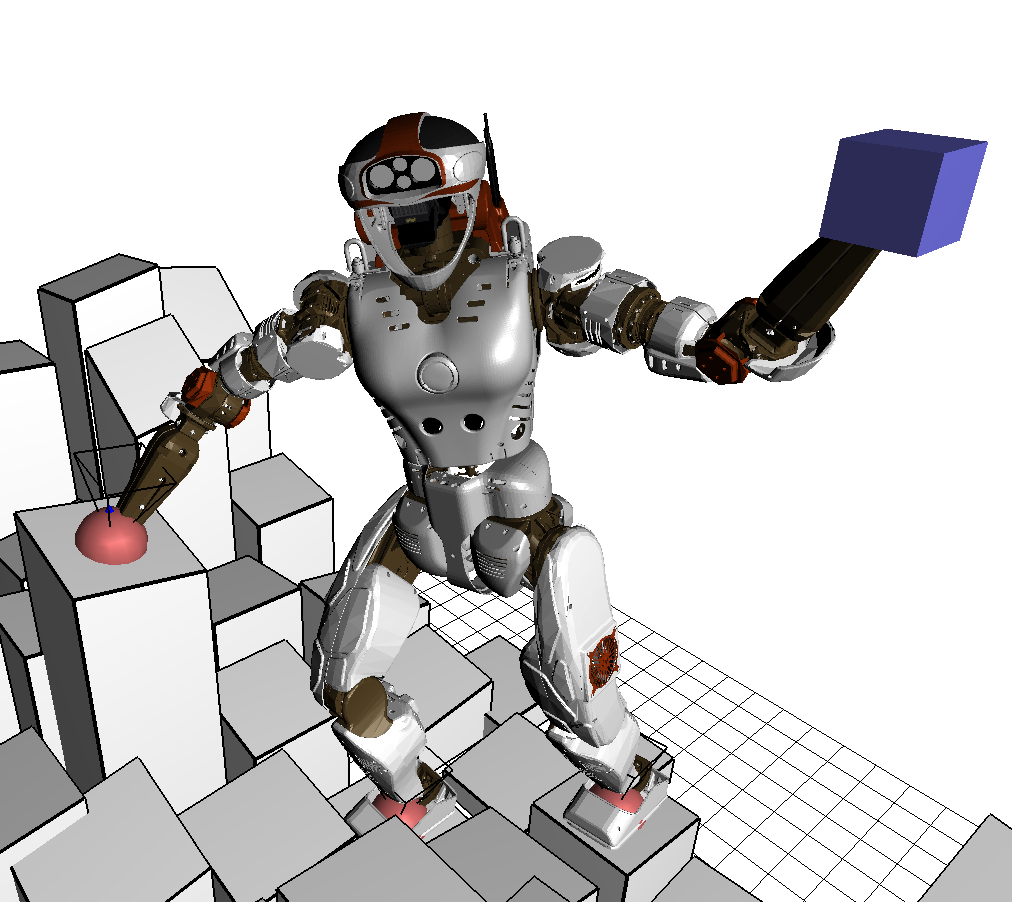}	
		\caption{Valkyrie}
	\end{subfigure}
	\begin{subfigure}{0.58\linewidth}
		\centering
		\includegraphics[height=3.5cm]{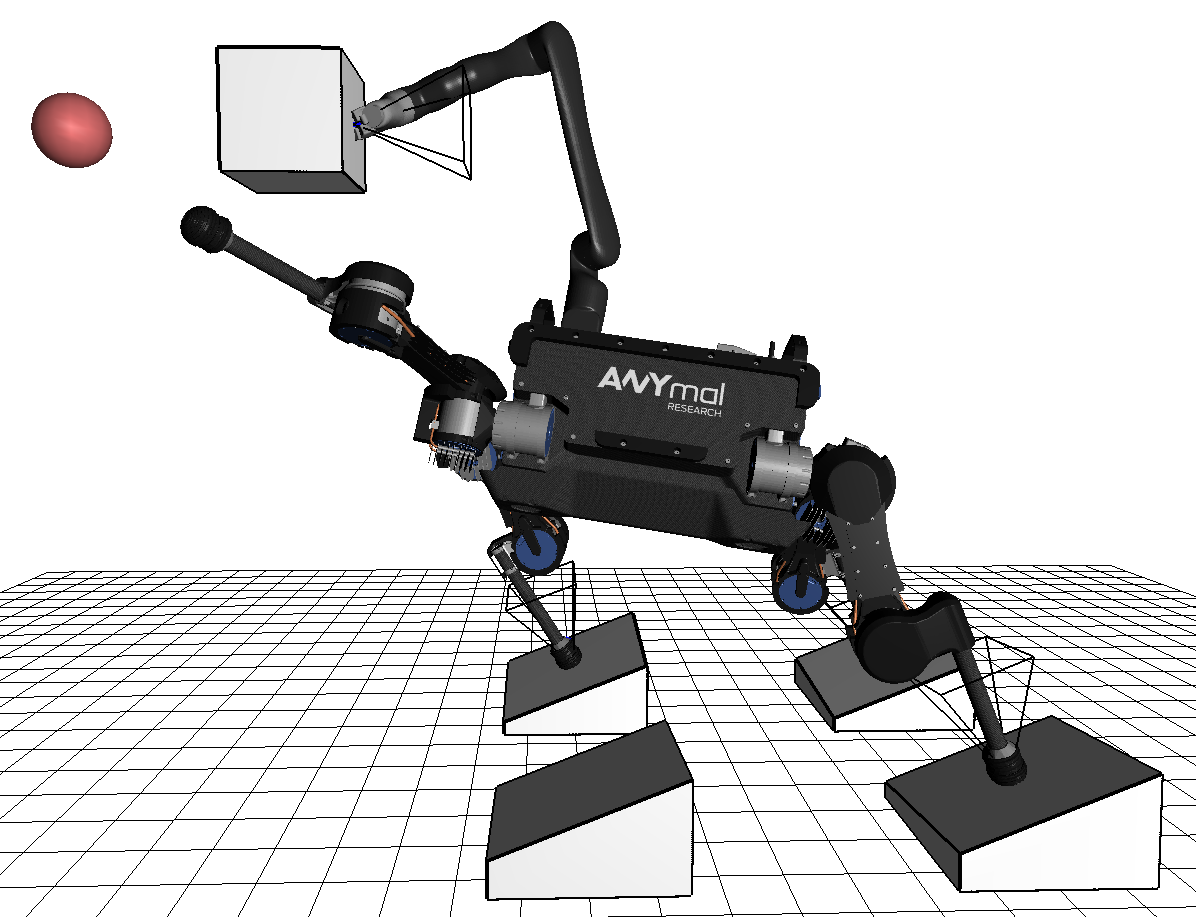}	
		\caption{ANYmal with a robot arm}
	\end{subfigure}
    \caption{
        Teleoperation of multi-contact interactions -- locomotion and manipulation on uneven terrains for humanoid and quadruped robots. % feasibility guaranteed
    }
    \label{fig:robot_platforms}
    \vspace{-4mm}
\end{figure}

%Problem of teleop complex system that can fall
The development of robotic teleoperation is advancing towards improving their versatile capabilities with complex platforms. For example, legged robots with manipulators and arms can be teleoperated to perform loco-manipulation tasks in challenging, unstructured, and natural terrains. However, legged robots, e.g. humanoids and quadrupeds, have a high number of degrees of freedom, making it difficult for operators to command all these joints directly while satisfying the balancing criteria. Besides, robotic systems are also subject to physical constraints, such as joint limits, actuator power limits and non slipping contacts, and the teleoperation system must consider all of these to ensure safety. When human operators make mistakes, the system should be robust and able to deal with any dangerous or infeasible commands.

To address these problems, the teleoperation system needs to retarget and adapt desired motions into a specific robot's morphology, while considering its physical limitations. While the balance on flat terrain can be analyzed by simple geometric criteria, such as the projection of the Center of Mass (CoM), the contact wrenches and force distribution need to be considered in complex non-coplanar multi-contact cases. Unlike the planning problem with known future states \cite{9145664}, online teleoperation system is an interactive scheme where only the current command of the operator is known. Therefore, the retargeting method must adapt the operator's input reactively to enforce safety in real time.

In this work, we developed a novel formulation to solve the motion retargeting as an optimization problem efficiently, with which operator's commands can be adapted and then executed on robots in real time -- guaranteeing all the feasibility constraints, and the balance of floating base robots as well.

The automatic adaptation and enforcement, including balance and other hard physical constraints, alleviate human mental load and allow the operator to focus on high-level supervision for solving complex loco-manipulation tasks -- shared control where humans provide task-level skills, and the algorithms resolve the local control of the high degree-of-freedom (DoF) robots and their physical constraints. Our method is suited for difficult high-DoF teleoperation where safety, feasibility, and effective prevention of erroneous commands are crucial.

\subsection{Related Works}\label{sec:biblio}

%Main story (teleoperation papers)
Conventional retargeting schemes for teleoperation use Inverse Kinematics (IK) extensively to compute whole-body joint positions from desired end-effectors and Center of Mass (CoM) references. 
For humanoids on flat ground \cite{MontecilloPuente2010OnRW,koenemann2014real,darvish2019whole,abi2018humanoid,9117048}, quasi-static equilibrium is formulated by constraining the CoM projection within the support polygon.

%IK not enough
However, IK-based schemes only consider kinematics constraints, multi-contact loco-manipulation on uneven surfaces require the use of force-related quantities to guarantee feasibility.
{OpenSoT} proposed an IK formulation to constrain both kinematic and dynamic quantities by integrating the joint accelerations at the velocity level \cite{hoffman2018robot}, but the contact wrenches are neither constrained nor optimized and must be provided as inputs additionally.

%Works have been done on teleop dynamic motions but only on flat ground
The balance of dynamic motions was studied on flat ground using the Zero Moment Point (ZMP) \cite{vukobratovic2004zero} and the Linear Inverted Pendulum Model (LIPM) \cite{kajita20013d},
based on which the dynamic filter \cite{yamane2003dynamics} was proposed to generate balanced motions by transforming the ZMP references commanded by the operator.
Similarly, the Divergent Component of Motion (DCM) was applied to predict the evolution of the system for long-term balance \cite{ishiguro2017bipedal}.
In \cite{Penco2019Teleop}, human motions were captured online and transferred to a humanoid robot on flat ground.
The above methods used simplified models to avoid the computationally expensive nonlinear whole-body model, but are limited to flat (or coplanar) surfaces and do not address multi-contact cases. The complex centroidal model \cite{del2018zero} and energy state based model \cite{7139908} for fall prediction are promising alternatives.

The retargeting in \cite{di2016multi} extended humanoid teleoperation to the multi-contact case where the contact switching, as well as the constraints of kinematics, torque and contact were realized by an inverse dynamics Quadratic Programming (QP) controller.
However, only coplanar contact surfaces were considered and the balance criterion purely relied on the kinematics of the projected CoM, so the method cannot address unstructured uneven terrains.

%Other similar works in offline planning field
There are previous research in planning similar to our proposed scheme. 
The work in \cite{bretl2008testing} solved kinematics and force-related quantities in a constrained nonlinear Sequential Quadratic Programming (SQP), assuming the static equilibrium. A sequence of keyframe configurations were optimized in \cite{shigematsu2019generating}, contact stance poses were solved in \cite{bouyarmane2012humanoid}, and uneven multi-contact postures were computed by using analytical partial derivatives in \cite{brossette2018multicontact}. Compared to these works designed for offline planning, our proposed scheme is developed specifically for online real-time applications as an interactive teleoperation process. We developed novel techniques detailed in the following to enable fast real-time computation.

\subsection{Contribution}

%What method proposed to solve the problems
This work proposes an optimization-based motion retargeting to teleoperate robots and achieve physically feasible, safe and balanced multi-contact tasks -- \textit{Sequential Equilibrium and Inverse Kinematics Optimization (SEIKO)} -- applicable and suitable for combined locomotion and manipulation on uneven surfaces, where only quasi-static and/or slow-medium speeds are required, and safety and risk mitigation are more critical. 

Our contributions are summarized as follows:
\begin{enumerate}
    \item \textbf{A new algorithmic formulation of SEIKO (Section~\ref{sec:eq_formulation}--\ref{sec:eq_decomposition})} with real-time performance to optimize the whole robot configuration of joint positions, torques, and contact forces under strict feasibility constraints.
    \item \textbf{Smooth multi-contact switching algorithm (Section~\ref{sec:method_contact_switching})} for transitions in-between adding--removing new physical contacts using SEIKO.
    \item \textbf{An integrated motion retargeting teleoperation framework (Section~\ref{sec:controller}, \ref{sec:robustness_feedback})} for safe and robust interactive loco-manipulation tasks in multi-contact scenarios.
\end{enumerate}

%Significance of the proposed method
The proposed SEIKO is validated on floating-base robots (humanoid, quadruped) on various multi-contact tasks, and considering both plane and point contacts (see Fig.~\ref{fig:robot_platforms}). The framework has flexibility to use different low-level controllers, e.g. inverse dynamics \cite{feng2014optimization} or admittance control \cite{autonrobot2016} to track references of robot posture and contact forces for stabilization.

Inverse dynamic controllers are designed to track dynamic motions and guarantee instantaneous dynamic stability, but can not guarantee the long-term balance alone. They react aggressively at the edge of the feasibility boundary and eventually fail when the input reference is physically infeasible. Usually, high-level planners can take care of the feasibility by pre-computing viable trajectories offline (either quasi-static or dynamic), but this approach is not applicable to online teleoperation, because the future operator's commands are unknown and subject to any changes. Hence, our proposed motion retargetting serves as a safety layer for interactive online teleoperation in the context of multi-contact.

Compared to previous works based on IK, the formulated SEIKO includes both kinematics and contact forces. This allows to undertake a broader set of tasks such as the contact switching on uneven multi-contact surfaces, postural optimization for minimizing joint torques or pushing tasks (Section~\ref{sec:push})), while safely ensuring the balance equilibrium.

The remainder of this paper is organized as follows. 
The teleoperation scheme is detailed in Section~\ref{sec:scheme}.
The core algorithmic details of SEIKO for retargeting and the contact switching are formulated in Section~\ref{sec:method}.
The validation is presented, evaluated and analyzed in Section~\ref{sec:results} 
with loco-manipulation tasks demonstrated in simulations.
The limitations are discussed in Section~\ref{sec:discussion}. Finally, we concluded and suggested future work in Section~\ref{sec:conclusion}.

\section{Multi-contact Teleoperation Framework}\label{sec:scheme}

\subsection{Command Paradigm}\label{sec:interface}

\begin{figure}[t]
    \centering
    \includegraphics[width=1.0\linewidth]{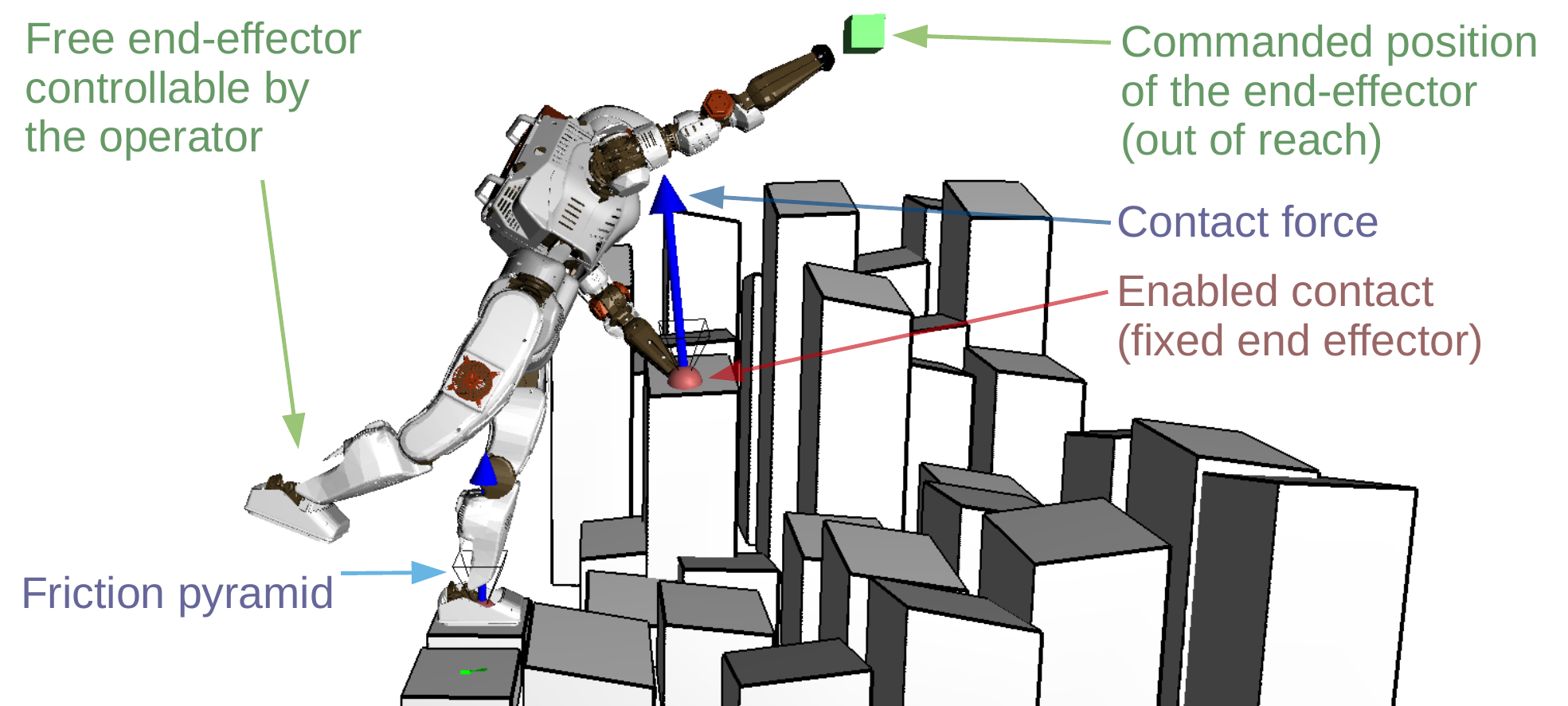}
    \caption{
        User interface for our multi-contact teleoperation, where an operator commands the pose of free end-effectors and can trigger contact switches.
    }
    \vspace{-4mm}
    \label{fig:val_end_effector}
\end{figure}

As depicted in Fig.~\ref{fig:val_end_effector}, the operator commands a high-DoF robot by mapping the end-effector motions to the whole-body configuration while guaranteeing feasibility, safety, and balance.
The robot establishes supporting contacts with the environment through the end-effectors, i.e. feet and hands. Each contact is categorized either as a planar (e.g., rectangular foot) or a point contact (e.g., hand stump), and has one contact state which is either \textit{enabled} or \textit{disabled} (i.e. free end-effector). 
The operator can provide three types of commands: the position and orientation for each free end-effector, a discreet contact switching trigger to remove or add a contact, and in an optional mode, a reference for the force applied by a specific contact.
The operator continuously commands the desired poses of the free end-effectors while the retargeting method optimizes the interaction forces at the enabled contacts.
At any time a contact switching smooth transition can be triggered (Section~\ref{sec:method_contact_switching}) to add or remove a selected end-effector.
The operator can also optionally define a desired normal contact force for a specific enabled contact to achieve pushing tasks.
Note that for point contacts, the surface orientation has to be externally provided.
As investigated in \cite{1525021}, the same formulation can allow the operator to command other predefined links of the robot such as the head, pelvis or shoulders. 

\begin{figure}[b]
    \centering
    \includegraphics[width=1.0\linewidth]{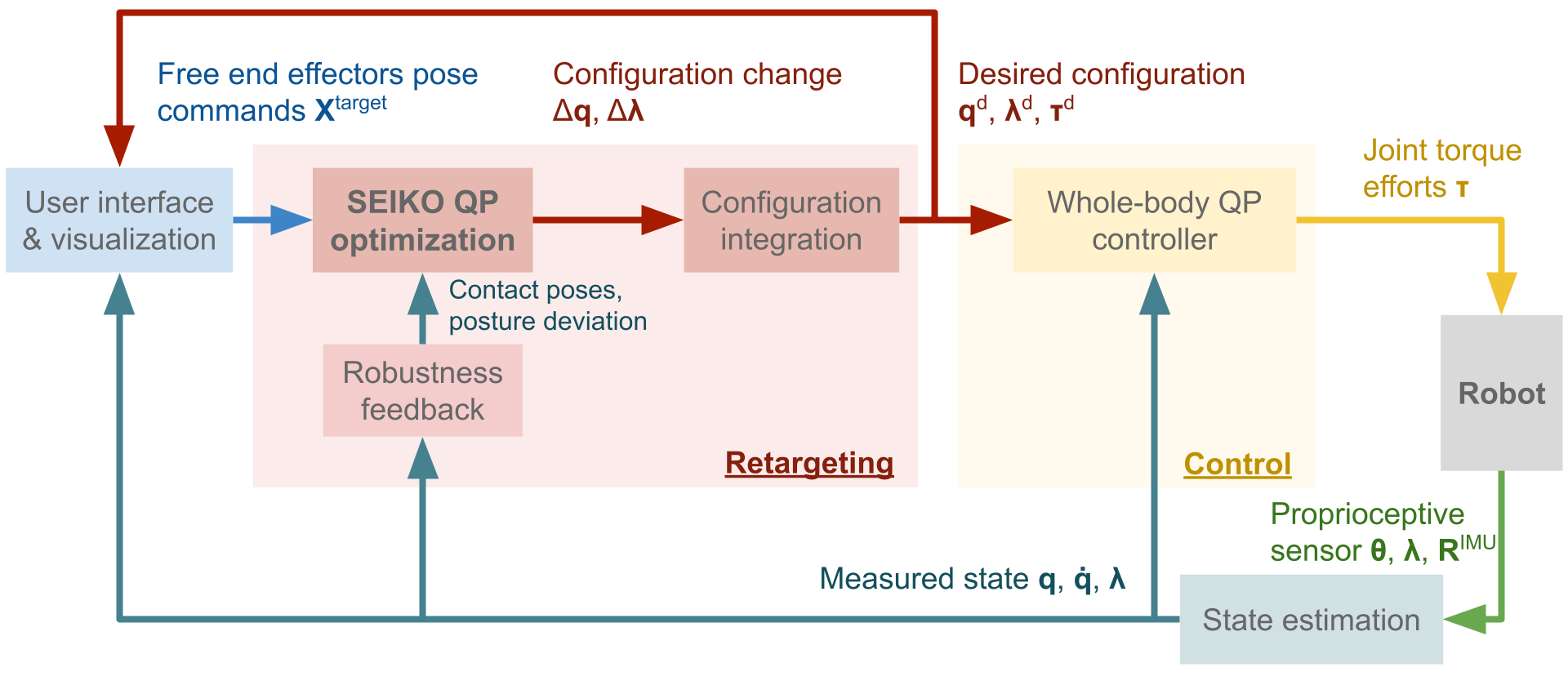}
    \caption{
        Two-stage retargeting and control architecture for multi-contact teleoperation.
        The operator commands the target poses $\bm{X}^{\text{target}}$ of the end-effectors. At each time step, SEIKO computes the incremental changes $\Delta \bm{q}, \Delta \bm{\lambda}$ which are integrated into the desired base and joint position $\bm{q}^d$, joint torque $\bm{\tau}^d$ and contact force $\bm{\lambda}^d$. Given the desired configuration, a whole body dynamic controller computes the joint torques $\bm{\tau}$ which are sent to the robot.
    }
    \label{fig:architecture}
    \vspace{-4mm}
\end{figure}
\subsection{Design of the Control Architecture}\label{sec:controller}

Fig.~\ref{fig:architecture} shows our 2-stage architecture of retargeting and control. The human operator continuously provides a high level Cartesian command $\bm{X}^{\text{target}}$ (pose of the end-effectors). In stage 1, the proposed SEIKO method retargets the commanded poses into best matching whole-body configuration. First an incremental configuration change [$\Delta\bm{q}$, $\Delta\bm{\lambda}$] is optimized with respect to all the physical constraints. The change is then integrated to produce a feasible desired configuration [$\bm{q}^{\text{d}}$, $\bm{\lambda}^{\text{d}}$, $\bm{\tau}^{\text{d}}$]. In stage 2, this desired configuration is tracked by a whole-body dynamic controller based on inverse dynamics that solves a Quadratic Programming (QP) \cite{feng2014optimization}. The controller solves and computes joint accelerations, joint torques and contact wrenches while optimizing a set of weighted tasks such as the positions of joints, CoM and contact forces. The joint torque efforts are then sent to the robot system. In supplementary materials, we discuss in Section~9 the limitations of QP controllers, and we show in Section~10 how essential it is to enforce the feasibility of input references.

\section{Principles and Formulation of SEIKO}\label{sec:method}

The physics of the robot is governed by the nonlinear equation of motion \cite{featherstone2014rigid}:
\begin{equation}
    \bm{M}(\bm{q})\bmdd{q} + \bm{C}(\bm{q},\bmd{q}) + \bm{G}(\bm{q}) = \bm{S}\bm{\tau} + \bm{J}(\bm{q})^\T\bm{\lambda},
\end{equation}
where $\bm{M}$ is the inertia matrix, $\bm{C}$ is the vector of centrifugal and Coriolis forces, $\bm{G}$ is the gravitational vector, $\bm{S}$ is the selection matrix for the underactuated floating base, $\bm{q}$ is the vector of generalized degrees of freedom positions including the pose of the floating base and joint positions (denoted as $\bm{\theta}$), $\bm{\tau}$ is the joint torques, $\bm{J}$ is the stacked Jacobian matrices of all contact points and $\bm{\lambda}$ is the stacked contact wrenches.

By limiting to slow and continuous commanded motions, we handle the unknown intention of the operator with one step ahead online optimization. With restrictions to quasi-static motions such that $\bmdd{q} \approx \bmd{q} \approx \bm{0}$, the terms related with acceleration and centrifugal and Coriolis forces become zero.

Classical dynamic QP controllers solve for the decision variable $\mat{\bmdd{q}, \bm{\tau}, \bm{\lambda}}^\T$. Hence, if the QP controllers as in \cite{feng2014optimization}, \cite{mansard2009versatile}, \cite{rocchi2015opensot} implement inequality constraints, they often only operate within a conservative subspace and remain away from their feasibility boundaries.

On the contrary, our formulations particularly address the requirements from multi-contact teleoperation, where stable configuration needs to be planned near and on the edge of the feasibility boundaries. This provides more possibilities to the operator to safely reach and operate at the boundaries, compared to the over restriction in the conventional formulation. We used an active set algorithm for solving the QP which has better numerical stability than interior point methods at the boundaries of inequality constraints.

\subsection{Optimization Formulation}\label{sec:eq_formulation}

The posture retargeting is formulated as constrained nonlinear optimization, which is solved by a sequence of QP problems. As operator's commands are constantly changing in real time, the problem is continuously updated at each time step. As our teleoperation use case requires interactive and reactive control, fast computation speed is critical. 

The proposed SEIKO differs from the classic SQP in three aspects. Firstly, we only compute one SQP iteration (one linearization and QP solution) per control loop. This allows online execution with a fast update frequency, i.e. $\SI{1}{\kilo\hertz}$. Second, the problem is updated at each control loop with the continuously changing commands of the operator. Third, classic SQP schemes use line search \cite{gill2012sequential} to improve convergence speed, i.e. the scalar step length is optimized in the gradient direction to minimize the cost function. However, in our case, our study found that line search increases computational time and is not needed, because the converged errors are sufficiently small (see Section~\ref{sec:result_convergence}). Therefore, our formulation keeps the SQP step length constant and equal to $1$, as this is the best \cite{kraft1988software} when the configuration is close to the optimal solution, which is our case because the problem (thus the optimal solution) changes slowly when continuously updated at a high frequency, and the initial configuration is always initialized from the measured robot state.\\

The SEIKO's QP problem is formulated as a constrained least square optimization that is solved at each control loop:
\begin{equation}
\begin{aligned}
    & \underset{\Delta\bm{x}}{\text{min}}~\norm{\bm{C}_{\text{cost}}(\bm{x}) \Delta\bm{x} -  \bm{c}_{\text{cost}}(\bm{x})}^2_{\bm{w}} \text{~~s.t.} \\
    & \bm{C}_{\text{eq}}(\bm{x}) \Delta\bm{x} + \bm{c}_{\text{eq}}(\bm{x}) = \bm{0}, \\
    & \bm{C}_{\text{ineq}}(\bm{x}) \Delta\bm{x} + \bm{c}_{\text{ineq}}(\bm{x}) \geqslant \bm{0}, \\
    & \text{where } \bm{x} = \mat{\bm{q}^{\text{d}} \\ \bm{\tau}^{\text{d}} \\ \bm{\lambda}^{\text{d}}},
    \Delta\bm{x} = \mat{\Delta\bm{q} \\ \Delta\bm{\tau} \\ \Delta\bm{\lambda}}. \\
\end{aligned}
\end{equation}

Here, $\bm{x}$ is the current desired configuration, and the incremental change $\Delta\bm{x}$ is the decision variable.
$\bm{C}_{\text{cost}},\bm{c}_{\text{cost}},\bm{C}_{\text{eq}},\bm{c}_{\text{eq}},\bm{C}_{\text{ineq}},\bm{c}_{\text{ineq}}$ are the matrices and vectors defining the cost, equality and inequality constraints respectively. Section~\ref{sec:eq_cost} to \ref{sec:eq_cst} describe the tasks and constraints which are stacked to define $\bm{C}_{\text{cost}},\bm{c}_{\text{cost}},\bm{C}_{\text{eq}},\bm{c}_{\text{eq}},\bm{C}_{\text{ineq}},\bm{c}_{\text{ineq}}$ and provided as input to the QP solver.

The motion equation under the quasi-static assumption is linearized and approximated at the first-order, and the analytical derivatives are used for better computation speed and stability. The quadratic cost function and linear constraints are thus formulated with respect to the decision variables.

In contrast to the usual QP formulation, our decision variable here is the incremental change $\mat{\Delta\bm{q}, \Delta\bm{\tau}, \Delta\bm{\lambda}}^\T$, which is equivalent as optimizing the rate change of the configuration $\mat{\bm{q}^{\text{d}}, \bm{\tau}^{\text{d}}, \bm{\lambda}^{\text{d}}}^\T$. The resulting configuration is then updated as:
\begin{align}
    \bm{x}_{t+1} = \bm{x}_{t} + \Delta\bm{x}.
\end{align}

In the proposed formulation, each new solution is a full configuration set which includes joint positions, joint torques, and contract wrenches $\mat{\bm{q}^{\text{d}}, \bm{\tau}^{\text{d}}, \bm{\lambda}^{\text{d}}}^\T$ at a \textit{stable static equilibrium}. The QP is guaranteed to have a solution because the solution $\mat{\bm{0}, \bm{0}, \bm{0}}^\T$ of no configuration changes is always valid. This satisfies the particular requirement for safety-critical teleoperation tasks where the system states always need to be stable so that the robot can halt instantly in case of emergency. The quasi-static motion allows the safe emergency stop at any time or when the feasibility boundary is reached.

The following sections describe the weighted cost function and constraints of the optimization (see detailed notations in supplementary materials Section~1). Note that the expressions of spatial algebra are simplified and the formal Lie algebra operations are in the supplementary materials. In the following sections, the desired configurations being optimized are denoted as $\bm{\theta},\bm{q},\bm{\tau},\bm{\lambda}$, instead of $\bm{\theta}^d,\bm{q}^d,\bm{\tau}^d,\bm{\lambda}^d$ for clarity.

\subsection{Optimization Formulation}\label{sec:eq_cost}

The optimization aims to minimize the weighted tasks:
\begin{equation}\label{eq:cost}
\begin{aligned}
    \text{min}~ &
    \norm{\dot{\bm{\theta}}}^2_{\bm{w}_{\text{velocity}}} + \norm{\bm{\tau}}^2_{\bm{w}_{\text{torque}}} +
    \sum_i \norm{\bm{\lambda}_i^{\text{target}}-\bm{\lambda}_i}^2_{\bm{w}_{\text{contact,~i}}} +\\
    & \norm{\mathsf{Clamp}\left(\bm{\theta}^{\text{target}}-\bm{\theta}\right)}^2_{\bm{w}_{\text{posture}}} +\\
    & \sum_i \norm{\mathsf{ClampNorm}\left(\bm{X}_i^{\text{target}} \ominus \bm{X}_i(\bm{q}) \right)}^2_{\bm{w}_{\text{position,i}},\bm{w}_{\text{orientation,~i}}}
\end{aligned}
\end{equation}

The joint velocities $\norm{\dot{\bm{\theta}}}^2_{\bm{w}_{\text{velocity}}}$ are minimized to enforce the quasi-static motions and to improve the optimization stability around kinematic singularities and feasibility boundaries.

The joint positions $\norm{\mathsf{Clamp}\left(\bm{\theta}^{\text{target}}-\bm{\theta}\right)}^2_{\bm{w}_{\text{posture}}}$ are attracted toward a default posture to regularize the nullspace of the end-effector's pose. This term typically helps the end-effector to recover its nominal pose after undergoing a highly singular motion. $\bm{\theta}^{\text{target}} \in \R^n$ is a nominal joint position vector.

The pose of the $i$th free end-effector $\norm{\mathsf{ClampNorm}\left(\bm{X}_i^{\text{target}} \ominus \bm{X}_i(\bm{q}) \right)}^2_{\bm{w}_{\text{position,i}},\bm{w}_{\text{orientation,~i}}}$ is driven towards a target pose in the Cartesian space,
where
$\bm{X}_i(\bm{q}) \in \SE$ is the current Cartesian pose measured by forward kinematics and
$\bm{X}_i^{\text{target}} \in \SE$ is the target Cartesian pose in world frame. 
The distance vectors in the joint and Cartesian space are clamped to prevent unbounded numerical values in the QP solver, which improves the stability of solving the optimization when the robot is operating or stuck at the feasibility boundaries. 
The clamping function $\mathsf{Clamp}()$ thresholds the absolute value of all individual input vector components, while the function $\mathsf{ClampNorm}()$ bounds the norm of the input vector in $\R^3$.

The orientations of enabled contact points are not constrained. We regulate the orientations towards the surface normals to avoid unexpected collisions with the environment.
The joint torques $\norm{\bm{\tau}}^2_{\bm{w}_{\text{torque}}}$ are minimized so that he optimized posture is regularized toward an energy efficient configuration.

The wrenches and forces of every enabled plane and point contacts $\norm{\bm{\lambda}_i^{\text{target}}-\bm{\lambda}_i}^2_{\bm{w}_{\text{contact,~i}}}$ are optimized respectively as close to a target wrench/force $\bm{\lambda}_i^{\text{target}} \in \R^6$/ $\R^3$ as possible. The target $\bm{\lambda}_i^{\text{target}}$ is $\bm{0}$ for an idle, non-contact end-effector, and it can be used to generate a desired contact force or a center of pressure (CoP).
The weights for each contact associated to the task $\bm{w}_{\text{contact,~i}}$ are used to regulate the force distribution among the contacts. The parameters of the cost function used in the experiments of the humanoid robot are listed in \autoref{table:ikid_parameters}. From extensive tests, these parameters were robust to various teleoperation tasks and requires no additional fine tuning to be transferred between two robots.

\begin{table}[t]
    \centering
    \caption{Typical parameters used during the humanoid and quadruped experiments. ($\mathsf{1}_n$ is the vector of ones of size the number of joint $n$)}
    \label{table:ikid_parameters}
    \begin{tabular}{|l|c|}
        \hline
        \textbf{Parameter} & \textbf{Value} \\
        \hline
        $\bm{w}_{\text{velocity}} = w_{\text{velocity}} \mathsf{1}_n$ & $10^4$ \\
        $\bm{w}_{\text{posture}} = w_{\text{posture}} \mathsf{1}_n$ & $1$ \\
        $\bm{w}_{\text{position}} = w_{\text{position}} \mathsf{1}_3$ & $10^3$ \\
        $\bm{w}_{\text{orientation}} = w_{\text{orientation}} \mathsf{1}_3$ & $1$~--~$10^2$ \\
        $\bm{w}_{\text{torque}} = w_{\text{torque}} \mathsf{1}_n$ & $10^{-5}$ \\
        $\bm{w}_{\text{contact}} = w_{\text{contact}} \mat{1 & 1 & 1 & 1 & 1 & 0.01}^\T$ & $10^{-5}$~--~$1$ \\
        $w^{\text{enabled}}_{\text{contact}}$ & $10^{-5}$ \\
        $w^{\text{disabled}}_{\text{contact}}$ & $1$ \\
        clamp bound for joint angular position & $0.1$~rad \\
        clamp bound for Cartesian position & $0.01$~m \\
        clamp bound for Cartesian orientation & $0.1$~rad \\
        $\alpha$ (contact switching transition factor) & $1.005$ \\
        \hline
    \end{tabular}
\end{table}

\subsection{Optimization Constraints}\label{sec:eq_cst}

Several types of constraints that assure the feasibility of the configuration are: kinematic constraints (joint position and velocity limits), actuator power constraints (joint torque limits) and balance constraints. The balance constraints require two conditions in the quasi-static case: first, all the external forces acting on the robot and the joint torques must follow the equilibrium equation; second, each contact must be stable, i.e. not slipping, not tilting and not pulling from the surface. 

All these constraints are formulated as linear equality and inequality constraints with respect to the decision variables, as in \cite{caron2015stability}. By constraining the system configuration within these bounds, we can guarantee that the robot is statically balanced with physically feasible postures while satisfying actuation requirements.\\

The equation of motion enforced at the static balancing equilibrium reduces to:
\begin{equation}\label{eq:equilibrium_equation}
    \bm{G}(\bm{q}) = \bm{S} \bm{\tau} + \bm{J}(\bm{q})^\T \bm{\lambda}.
\end{equation}

In the world frame, the pose of the $i$th contact points and planes is defined with the kinematic constraints:
\begin{equation}\label{eq:cst_kinematic}
\begin{aligned}
    & \bm{X}_i^{\text{target}} \ominus \bm{X}_i(\bm{q}) = \bm{0} & ~\text{for plane contacts}, \\
    & \bm{p}_i^{\text{target}} - \bm{p}_i(\bm{q}) = \bm{0} & ~\text{for point contacts},
\end{aligned}
\end{equation}
where $\bm{X}_i^{\text{target}} \in \SE$ and $\bm{p}_i^{\text{target}} \in \R^3$ are the pose and position measured from the robot's state when the $i$th contact is established.\\

The classical inequality constraints (detailed in supplementary materials Section~2 and in \cite{caron2015stability}) enforce the joint position and torque limits of the system and enable feasible contact conditions with the constrained normal force, center of pressure, friction pyramid and torsional torque.

\subsection{Partial Derivatives}\label{sec:eq_partial}

The cost, equality and inequality constraints of the optimization in Section~\ref{sec:eq_cost}, Section~\ref{sec:eq_cst} are written in the nonlinear form for clarity. However, solving the QP at each control loop for the posture change $\Delta\bm{x}$, requires the expressions of their first order differentiation. Most of the differentiated terms are simple. We focus on $\mathsf{ClampNorm}\left(\bm{X}_i^{\text{target}} \ominus \bm{X}_i(\bm{q}) \right)$ from \eqref{eq:cost} and \eqref{eq:equilibrium_equation} and \eqref{eq:cst_kinematic} which are non-trivial.

The target cost for free end-effector pose $i$ in \eqref{eq:cost} is differentiated as
\begin{equation}
    \mathsf{ClampNorm}\left(\bm{X}_i^{\text{target}} \ominus \bm{X}_i(\bm{q})\right) + \bm{J}_i(\bm{q}) \Delta\bm{q},
\end{equation}
and the plane contact $i$ in \eqref{eq:cst_kinematic} for kinematics constraint yields: 
\begin{equation}
    \bm{X}_i^{\text{target}} \ominus \bm{X}_i(\bm{q}) + \bm{J}_i(\bm{q}) \Delta\bm{q} = \bm{0},
\end{equation}
where $\bm{J}_i(\bm{q}) \in \R^{6 \times (6+n)}$ is the Jacobian of the end-effector frame $i$ expressed in world frame. For point contacts, only the linear part is used.

The equilibrium in \eqref{eq:equilibrium_equation} is nonlinear in the gravitational term. Using the incremental change in the configuration $\Delta\bm{x}$, we differentiate the equilibrium equation as:
\begin{equation}\label{eq:diff_equilibrium}
    \bm{G}(\bm{q}+\Delta \bm{q}) = \bm{S} (\bm{\tau}+\Delta\bm{\tau}) + \bm{J}(\bm{q}+\Delta \bm{q})^\T (\bm{\lambda} + \Delta\bm{\lambda}).
\end{equation}
% start a new line for a new paragraph 

If \eqref{eq:equilibrium_equation} is differentiated only by $\Delta\bm{q}$, the term $\left(\fracdiff{\bm{J}}{\bm{q}} \Delta\bm{q}\right)^T\bm{\lambda}$ will appear which is bilinear in $(\Delta\bm{q},\bm{\lambda})$ and cannot be expressed by the formulation of linear equality constraint of the QP. Therefore, \eqref{eq:equilibrium_equation} is differentiated by $\Delta\bm{q}, \Delta\bm{\tau}, \Delta\bm{\lambda}$. By using the first-order terms, \eqref{eq:diff_equilibrium} is approximated as:
\begin{align}
    \bm{G}(\bm{q}) +
    \fracdiff{\bm{G}}{\bm{q}} \Delta\bm{q} = 
    & \bm{S} \bm{\tau} + 
    \bm{S} \Delta\bm{\tau}  +\bm{J}(\bm{q})^T \bm{\lambda}\\
    & + \bm{J}(\bm{q})^T \Delta\bm{\lambda} + 
    \left(\fracdiff{\bm{J}}{\bm{q}} \Delta\bm{q}\right)^T \bm{\lambda}\nonumber
\end{align}
where $\fracdiff{\bm{G}}{\bm{q}}(\bm{q}) \in \R^{(6+n) \times (6+n)}$ is the partial derivatives of $\bm{G}(\bm{q})$
and $\fracdiff{\bm{J}}{\bm{q}}(\bm{q}) \in \R^{l \times (6+n) \times (6+n)}$ the kinematics Hessian tensor of the stacked Jacobian of contacts $\bm{J}(\bm{q})$.

The Hessian tensor product $\bm{H} \in \R^{(6+n) \times (6+n)}$ can be rewritten such as:
\begin{align}\label{eq:hessian_product1}
    &\left(\fracdiff{\bm{J}}{\bm{q}} \Delta\bm{q}\right)^T \bm{\lambda} = \bm{H} \Delta\bm{q},\\
\text{with} \quad
    & H_{ij} 
    = \sum_{k=1}^{l} \left(\fracdiff{\bm{J}}{\bm{q}}\right)_{kij} \lambda_k 
    = \left(\left(\fracdiff{\bm{J}}{q_j}\right)^\T \bm{\lambda}\right)_i. \label{eq:hessian_product2}
\end{align}

The differentiated equation of motion can then be linearly expressed
with respect to $\Delta\bm{x}$ as:
\begin{multline}\label{eq:full_diff_eq_motion}
    \mat{
        \left(\fracdiff{\bm{G}}{\bm{q}} - \bm{H}\right) &
        -\bm{S} &
        -\bm{J}(\bm{q})^\T
    }
    \Delta\bm{x} + \bm{G}(\bm{q}) \\
    - \bm{S} \bm{\tau} - \bm{J}(\bm{q})^T \bm{\lambda}
    = \bm{0}.
\end{multline}

\subsection{Decomposition of the Equation of Motion}\label{sec:eq_decomposition}

Utilizing the selection matrix, the equation of motion can be split into upper rows (floating base) and lower rows (joint space), as in \cite{herzog2016momentum}.
This decomposition allows to express the joint torques $\bm{\tau}$ linearly by the contact forces $\bm{\lambda}$, so the QP is solved much faster by removing joint torques from the decision variables. % you mean?: linearly from the contact forces --> linearly by the contact forces.
The same approach is applied to our differentiated equation of motion by slicing the $6$ floating base rows ($\mathsf{B}$) from the $n$ joints rows ($\mathsf{J}$):
\begin{equation}
\begin{aligned}
    & \bm{G} = \mat{\bm{G}_{\mathsf{B}} \\ \bm{G}_{\mathsf{J}}},
    \bm{J} = \mat{\bm{J}_{\mathsf{B}} & \bm{J}_{\mathsf{J}}},
    \fracdiff{\bm{G}}{\bm{q}} = \mat{\fracdiff{\bm{G}}{\bm{q}}_{\mathsf{B}} \\ \fracdiff{\bm{G}}{\bm{q}}_{\mathsf{J}}}
    \bm{H} = \mat{\bm{H}_{\mathsf{B}} \\ \bm{H}_{\mathsf{J}}} \\
    & \bm{G}_{\mathsf{B}} \in \R^6,
    \bm{G}_{\mathsf{J}} \in \R^n,
    \bm{J}_{\mathsf{B}} \in \R^{l \times 6},
    \bm{J}_{\mathsf{J}} \in \R^{l \times n}, \\
    & \fracdiff{\bm{G}}{\bm{q}}_{\mathsf{B}}, \bm{H}_{\mathsf{B}} \in \R^{6 \times (n+6)},
    \fracdiff{\bm{G}}{\bm{q}}_{\mathsf{J}}, \bm{H}_{\mathsf{J}} \in \R^{n \times (n+6)}.
\end{aligned}
\end{equation}

By applying such upper-lower partitions, the differentiated equation of motion \eqref{eq:full_diff_eq_motion} can be replaced by the two partitioned equations below:
\begin{equation}\label{eq:base_diff_eq_motion}
    \mat{
        \left(\fracdiff{\bm{G}}{\bm{q}}_{\mathsf{B}} - \bm{H}_{\mathsf{B}}\right) &
        -\bm{J}_{\mathsf{B}}(\bm{q})^\T
    }
    \mat{\Delta\bm{q} \\ \Delta\bm{\lambda}} 
    + \bm{G}_{\mathsf{B}}(\bm{q})
    - \bm{J}_{\mathsf{B}}(\bm{q})^T \bm{\lambda}
    = \bm{0},
\end{equation}
\begin{equation}\label{eq:joint_diff_eq_motion}
\begin{aligned}
    & \bm{\tau} + \Delta\bm{\tau} = \bm{T} \mat{\Delta\bm{q} \\ \Delta\bm{\lambda}} + \bm{t}, ~~\text{where}\\
    & \bm{T} = 
    \mat{
        \left(\fracdiff{\bm{G}}{\bm{q}}_{\mathsf{J}} - \bm{H}_{\mathsf{J}}\right) &
        -\bm{J}_{\mathsf{J}}(\bm{q})^\T
    }
    \in \R^{n \times (6+n+l)}, \\
    & \bm{t} = \bm{G}_{\mathsf{J}}(\bm{q}) - \bm{J}_{\mathsf{J}}(\bm{q})^T \bm{\lambda}
    \in \R^{n}.
\end{aligned}
\end{equation}
As shown in \eqref{eq:base_diff_eq_motion}, the QP only needs to optimize $\mat{\Delta\bm{q },  \Delta\bm{\lambda}}^T \in \R^{6+n+l}$ and the resulting joint torques 
$\bm{\tau} + \Delta\bm{\tau} \in \R^n$ can be equivalently and linearly expressed by \eqref{eq:joint_diff_eq_motion}.

\subsection{Contact Switching}\label{sec:method_contact_switching}

\begin{algorithm}[t]
    \footnotesize
    \caption{Disable contact $i$}
    \label{algo:contact_disable}
    \SetAlgoLined
    \DontPrintSemicolon
    $w_{\text{contact,~i}} \leftarrow w^{\text{enabled}}_{\text{contact}}$\tcp*[f]{Initial weighting value before switching}\;
    \While{$w_{\text{contact,~i}} < w^{\text{disabled}}_{\text{contact}}$}{
        $w_{\text{contact,~i}} \leftarrow \alpha w_{\text{contact,~i}}$ (with $\alpha > 1$)\;
        SEIKO online retargeting:
        $\begin{cases}$
        Solve $(\Delta\bm{q},\Delta\bm{\lambda}) \leftarrow \mathsf{SEIKO\_QP}(\bm{q}, \bm{\tau}, \bm{\lambda}, \bm{w})$\; $\\$
        Compute $\Delta \bm{\tau}$ from $(\Delta\bm{q},\Delta\bm{\lambda})$\; $\\$
        Integrate state $(\bm{q}, \bm{\tau}, \bm{\lambda}) \mathrel{+}\mathrel{\mkern-2mu}= (\Delta \bm{q}, \Delta \bm{\tau}, \Delta \bm{\lambda})$\;
        $\end{cases}$
    }
    \eIf{$\norm{\bm{\lambda}_i} < \epsilon$} {
        Disable contact $i$\;
        \Return Success\;
    }{
        Slowly decrease $w_{\text{contact,~i}}$ to come back to $w^{\text{enabled}}_{\text{contact}}$\;
        \Return Failure\;
    }
\end{algorithm}

The ability to add and remove contacts during teleoperation allows a broader range of manipulation and locomotion tasks, but requires a smooth transition and enforced feasibility constraints. Both the force distribution and the kinematic posture have to change, in order to free a contact point, which cannot be achieved by a pure IK formulation.

Removing a contact needs to smoothly bring contact forces to zero, see the procedure in \autoref{algo:contact_disable}.
The transition is implemented by exponentially increasing the penalty weight
associated to the contact force regularization $w_{\text{contact,~i}}$ from $w^{\text{enabled}}_{\text{contact}}$ to $w^{\text{disabled}}_{\text{contact}}$ (see parameters in \autoref{table:ikid_parameters}). The duration of this transition is defined by the transition factor $\alpha$ and the update frequency. In our tests, this procedure runs online, and the parameter $w_{\text{contact,~i}}$ is being changed while SEIKO optimization keeps running continuously.

Our formulation, which combines both kinematics and force quantities, naturally shifts the posture and force distribution toward the remaining contacts when the wrench penalty on a specific contact is increased. This transition motion is induced by optimizing the regularization terms $\norm{\bm{\tau}}^2_{\bm{w}_{\text{torque}}}$,$\norm{\bm{\lambda}_i^{\text{target}}-\bm{\lambda}_i}^2_{\bm{w}_{\text{contact,~i}}}$ in \eqref{eq:cost} and the equilibrium equality constraint \eqref{eq:equilibrium_equation}.

To add a new contact, this procedure simply needs to run reversely by changing from $w_{\text{contact}} = w^{\text{disabled}}_{\text{contact}}$ to $w^{\text{enabled}}_{\text{contact}}$, and the posture will change and the contact forces will smoothly redistribute. Note that removing a contact point by smoothly bring contact forces to zero is not always feasible. Such cases may occur when the inequality constraints prevent the posture and the force distribution from fully transferring to other supporting contacts, and the algorithm will fail to solve and remains at the initial contact state.

\subsection{Improvements of Robustness}\label{sec:robustness_feedback}

The two-stage architecture in Fig.~\ref{fig:architecture} consists of motion retargeting and control execution. First, the desired configuration is optimized by SEIKO; second, the measured configuration is estimated from the sensors and used by the dynamic controller to track the desired configuration. To provide a useful and relevant reference, the desired configuration must be consistent with the actual measured state of the robot. This consistency deteriorates when the pose of the contacts mismatch between these two configurations, for example in case of external pushes, slipping contact, or tracking errors. Hence, we take advantage of the online computation of SEIKO to formulate two feedback actions to improve the robustness of teleoperation. 

The measured pose of each enabled contact is estimated, filtered and used in the kinematic constraint \eqref{eq:cst_kinematic} of the desired configuration $\bm{X}^{\text{target}}_i=\bm{X}^{\text{measured}}_i$. Note that this does not generate drifting motion of end-effectors, because both SEIKO and the QP tracking controller assume fixed contacts.

In case of external pushes, the real posture of the robot can deviate from the desired one. We clamp the maximum angular distance between the desired and measured joint positions $\bm{q} = \bm{q}^{\text{measured}} + \mathsf{Clamp}(\bm{q}-\bm{q}^{\text{measured}})$. Within this angular range and thanks to the controller, the desired posture in joint space acts as a spring-damper attractor. When the angular distance becomes larger than the threshold, the desired posture follows the measured one and acts as a saturation. This feature is useful for safe physical interactions.

\subsection{Implementation}

Note that instead of building the costly full kinematic Hessian tensor \eqref{eq:hessian_product1} and \eqref{eq:hessian_product2}, only the Hessian-vector product is computed from the differentiation of the Recursive Newton-Euler Algorithm (RNEA) by setting $\bm{\ddot{q}} = \bm{\dot{q}} = \bm{0}$:
\begin{equation}
    \fracdiff{\mathsf{ID}}{\bm{q}} = 
    \fracdiff{\bm{M}}{\bm{q}}\bm{\ddot{q}} + 
    \fracdiff{\bm{C}}{\bm{q}}\bm{\dot{q}} + 
    \fracdiff{\bm{G}}{\bm{q}} - 
    \fracdiff{\bm{J}^\T}{\bm{q}}\bm{\lambda}.
\end{equation}

The proposed algorithms were implemented in C++ using \textit{RBDL} \cite{Felis2016} and \textit{Pinocchio} \cite{carpentier2019pinocchio} rigid-body libraries. \textit{Pinocchio} provides efficient and analytical computation of the partial derivatives of the equation of motion \cite{carpentier2018analytical}, hence the matrices $\fracdiff{\bm{G}}{\bm{q}}$ and $\fracdiff{\bm{J}^\T}{\bm{q}} \bm{\lambda}$ in \eqref{eq:full_diff_eq_motion} can be quickly retrieved. The QP solver uses \textit{EiQuadProg++} based on the algorithm in \cite{goldfarb1983numerically}.

Let $n$, $m_{\text{plane}}$ and $m_{\text{point}} \in \N$ be the number of joints, the numbers of currently enabled plane and point contacts respectively. At each time step, we solve a QP problem of $6+n+6m_{\text{plane}} + 3m_{\text{point}}$ decision variables. The total number of equality constraints is $m_{\text{eq}} = 6 + 6m_{\text{plane}} + 3m_{\text{point}}$.
Each plane contact generates $18$ inequality constraints while each point contact generates $6$. In total, the number of inequalities is $m_{\text{ineq}} = 2n + 2n + 18m_{\text{plane}} + 6m_{\text{point}}$.

\section{Results}\label{sec:results}

This section presents the validation results of our proposed teleoperation framework on two types of legged robots -- a humanoid and a quadruped -- performing complex multi-contact motions on uneven terrains (see the real-time performance in the attached video).
The retargeting capabilities were teleoperated online using real-time implementation with all feasibility constraints enforced.
The accompanying video of this paper summaries our approach and demonstrates all the validations of both robots during several teleoperated tasks.

\subsection{Computational Time}

\begin{table}[h!]
    \centering
    \caption{
        Average and maximum computing time for one control/optimization step
        (32 DOFs including the floating base).
    }
    \label{table:computing_time}
    \begin{tabular}{|l|p{1.3cm}|p{0.6cm}|}
        \hline
        \textbf{Task} & \textbf{Avg (max)\newline time (ms)} & \textbf{Ratio (\%)}\\
        \hline
        State estimation (filtering and model update) & 0.05 (0.09) & \\
        \hline
        Proposed SEIKO method & \textbf{0.47 (0.89)} & 100\%\\
        \hline
        -- Jacobian and gravity vector (\textit{RBDL})& 0.03 (0.05) & 6\%\\
        -- Analytical partial derivatives (\textit{Pinocchio})& 0.06 (0.13) & 12\%\\
        -- Cost, equalities and inequalities matrices & 0.11 (0.24) & 23\%\\
        -- QP solver & 0.16 (0.31) & 34\%\\
        -- Joint torques & 0.11 (0.19) & 23\%\\
        \hline
        Inverse dynamic QP controller & 0.28 (0.55) & \\
        \hline
        \textbf{Total control cycle} & 0.81 (1.48) & \\
        \hline
    \end{tabular}
\end{table}

The average and maximum computing times measured on an embedded mini-PC (Intel NUC, Intel Core i7-3615QE (2.30~GHz)) with a real-time Linux kernel for one control cycle is given in \autoref{table:computing_time}. The proposed method achieved good real-time performances thanks to the analytical partial derivatives and the decomposition of the motion equations \eqref{eq:base_diff_eq_motion} and \eqref{eq:joint_diff_eq_motion}. According to our extensive tests, the average computing time is fairly stable; however, the maximum time can vary depending on the kernel, CPU core binding, and other scheduling configurations of the operating system.

As previously mentioned in Section~\ref{sec:method_contact_switching}, the contact switching algorithm~\ref{algo:contact_disable} runs online and does not cause additional computing cost. Also, it can run offline for verification by computing only SEIKO's retargeting without executing the controller, to see if a contact switch is feasible without actually moving the robot. Assuming an update frequency of \SI{1000}{\hertz}, the transition requires $2309$ iterations to finish with the parameters listed in Table~\ref{table:ikid_parameters}. The transition duration is therefore $\SI{2.309}{\second}$, which can be computed offline in about $\SI{0.560}{\second}$ only, in the case of the Valkyrie robot.

\subsection{Convergence of the Equality Constraints}\label{sec:result_convergence}

\begin{figure}[H]
    \centering
    \begin{subfigure}{0.32\linewidth}
		\centering
		\includegraphics[width=\linewidth]{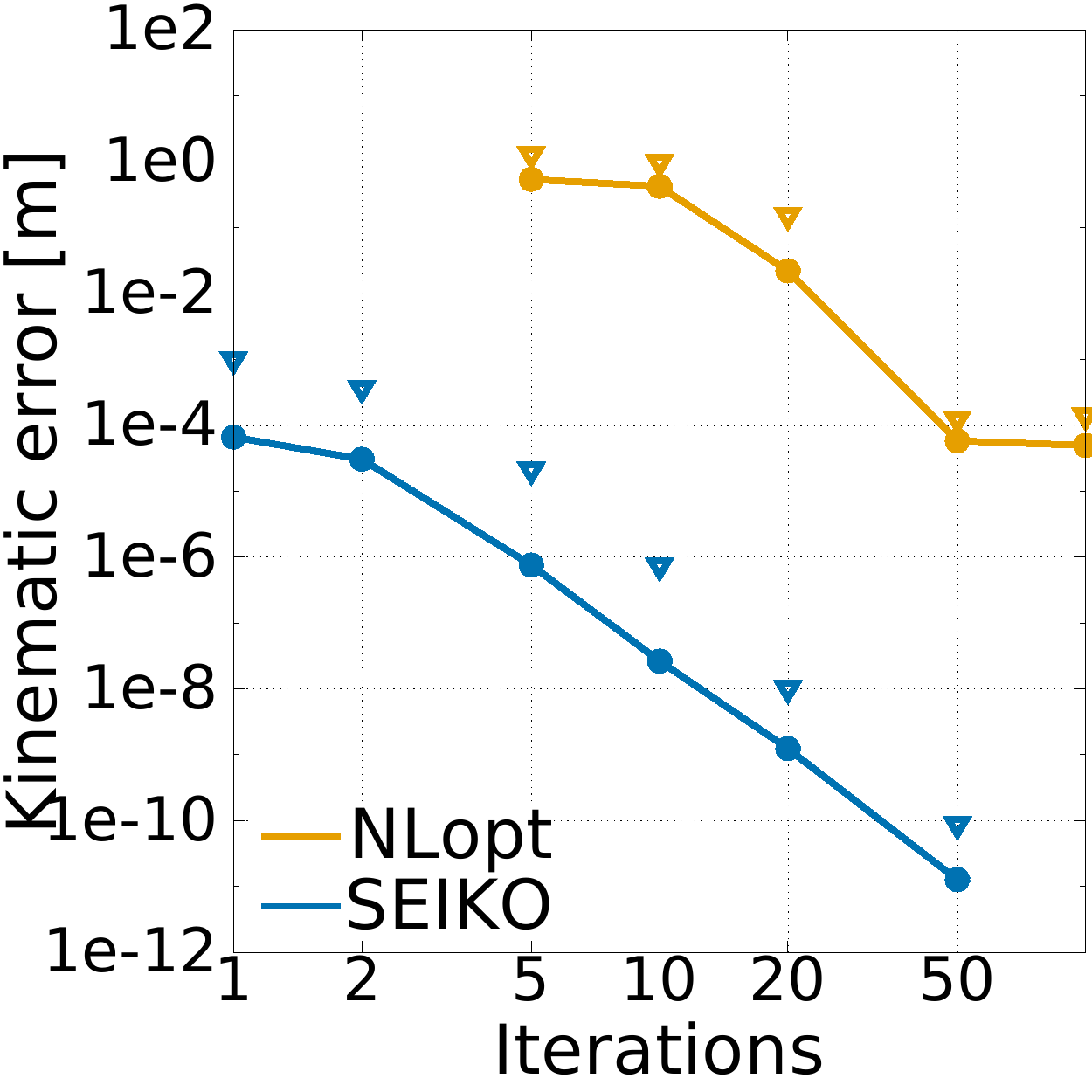}
		\caption{Kinematics}
	\end{subfigure}
	\begin{subfigure}{0.32\linewidth}
		\centering
		\includegraphics[width=\linewidth]{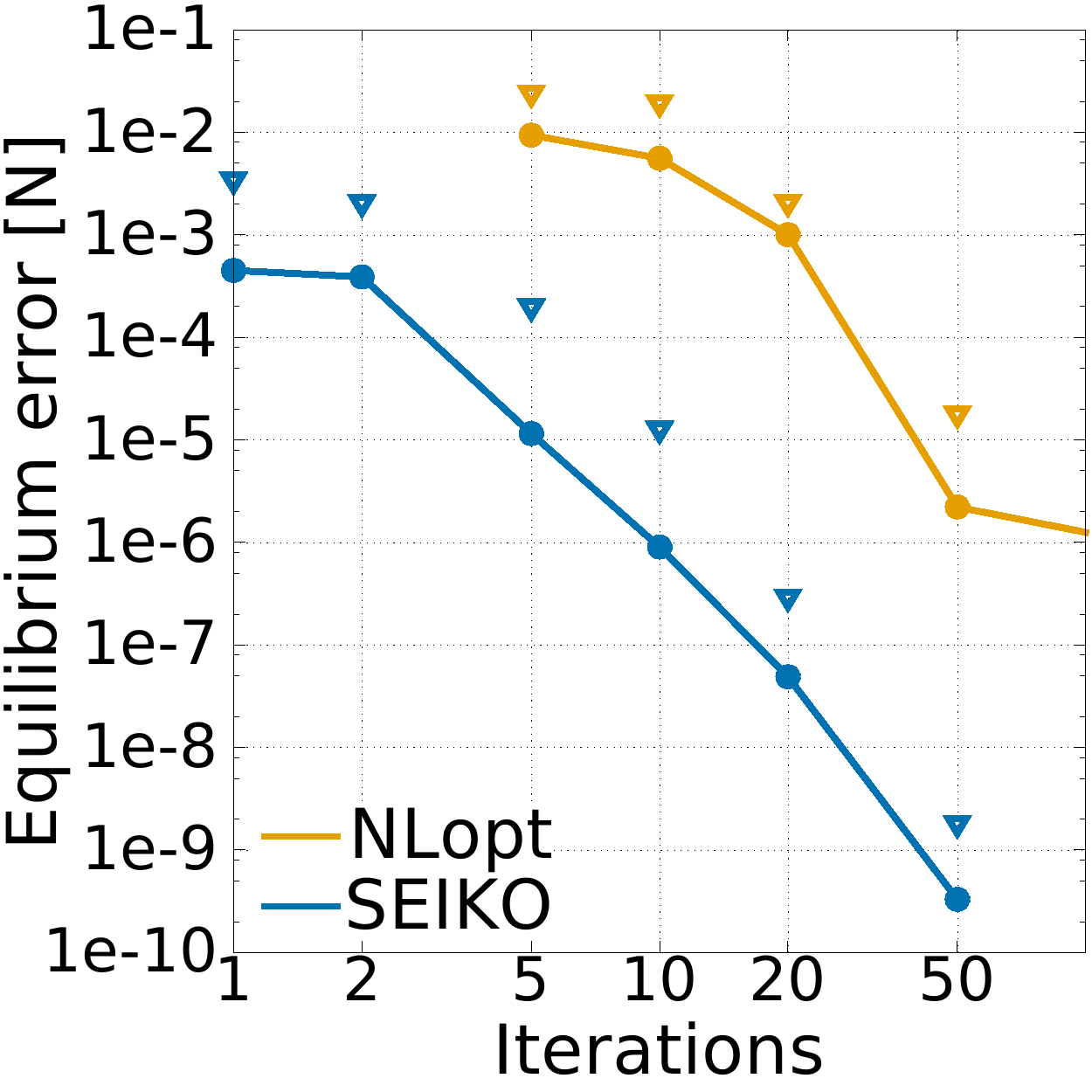}
		\caption{Equilibrium}
	\end{subfigure}
	    \begin{subfigure}{0.32\linewidth}
		\centering
		\includegraphics[width=\linewidth]{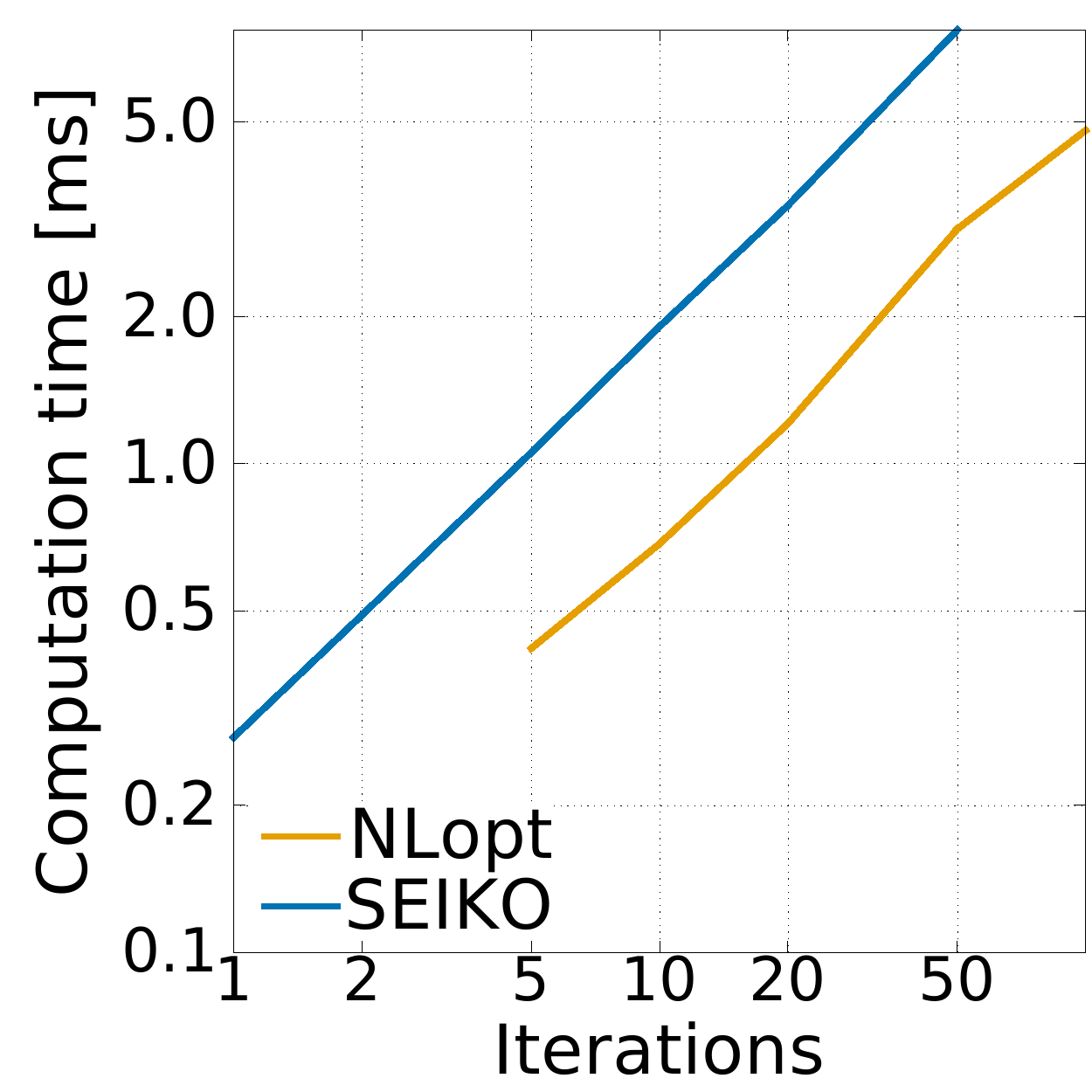}
		\caption{Computing time}
	\end{subfigure}
    \caption{
        Comparison of SEIKO and SLSQP NLopt by number of iterations: 
        The average (dots) and maximum errors (triangles) of the kinematic (a) and equilibrium (b) constraints; (c) average computational time per control loop.
    } \label{fig:plot_error_constraints}
\end{figure}

We compared SEIKO with the constrained nonlinear optimization SLSQP algorithm \cite{kraft1988software} provided by the NLopt library as a baseline. The SLSQPT was implemented with the same constraints (Section~\ref{sec:eq_cst}) and the same objectives (Section~\ref{sec:eq_cost}), but employs direct decision variables as the configuration of the system $\mat{\bm{q}^{\text{d}}, \bm{\tau}^{\text{d}}, \bm{\lambda}^{\text{d}}}^\T$, compared to our formulation of using increments.
Fig.~\ref{fig:plot_error_constraints} shows the residual errors of the kinematic \eqref{eq:cst_kinematic} and equilibrium \eqref{eq:equilibrium_equation} constraints as well as the computing time for one time step.
On the contrary to SEIKO, SLSQP relies on a line search algorithm to improve its convergence. Our tests showed that SLSQP requires multiple iterations to converge to satisfying constrain errors, where a single iteration does not produce a viable result.

In contrast, with only one iteration, our proposed SEIKO can achieve the position error below $\SI{1}{\milli\meter}$ for kinematics, and the force error below $\SI{0.01}{\newton}$ for the equilibrium constraints, which are all small enough to ensure accuracy for the teleoperation. These sub decimal scale of errors are especially negligible, compared to the uncertainties in the model and the controller.
It shall be noted that in some cases, SLSQP produces discontinuous successive solutions and thus jerky motions when the solution lies on the edge of the feasibility boundaries. The saturation of inequality constraints severely impedes the convergence of the SLSQP algorithm even with an increased number of iterations. More details are given on this phenomenon in supplementary materials in Section~7.

\subsection{Validation of Online Teleoperation}

\begin{figure*}[h!]
    \begin{minipage}{.16\textwidth}
        \begin{subfigure}{1.0\textwidth}
		    \centering
		    \includegraphics[height=2.2cm,trim={0 100 0 100},clip]{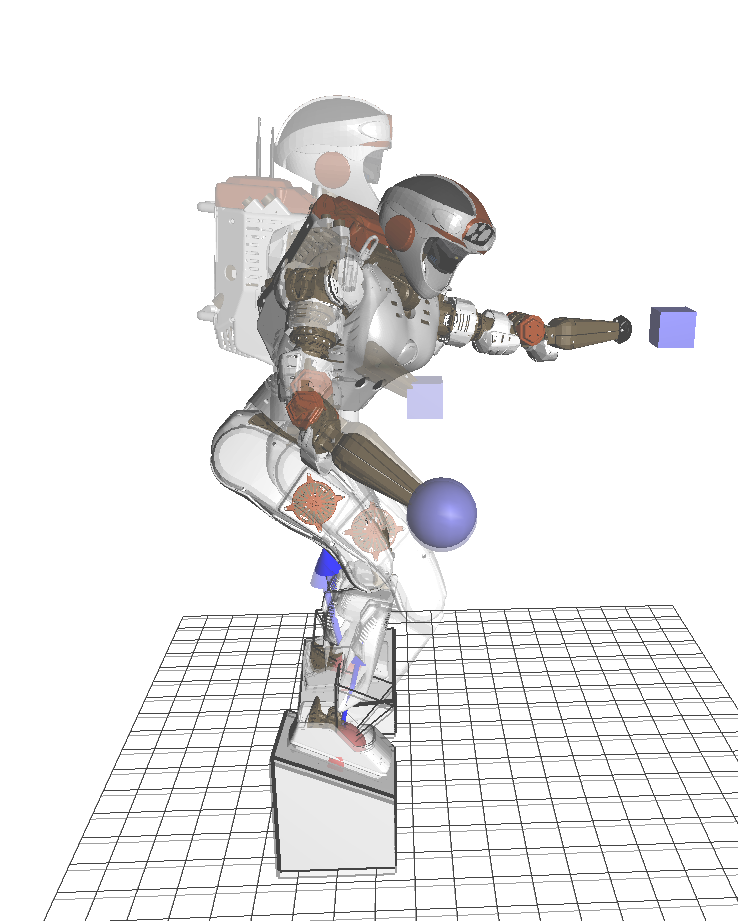}
		    \caption{Far reaching}
	    \end{subfigure}
	    \begin{subfigure}{1.0\textwidth}
		    \centering
		    \includegraphics[height=2.2cm,trim={0 100 0 100},clip]{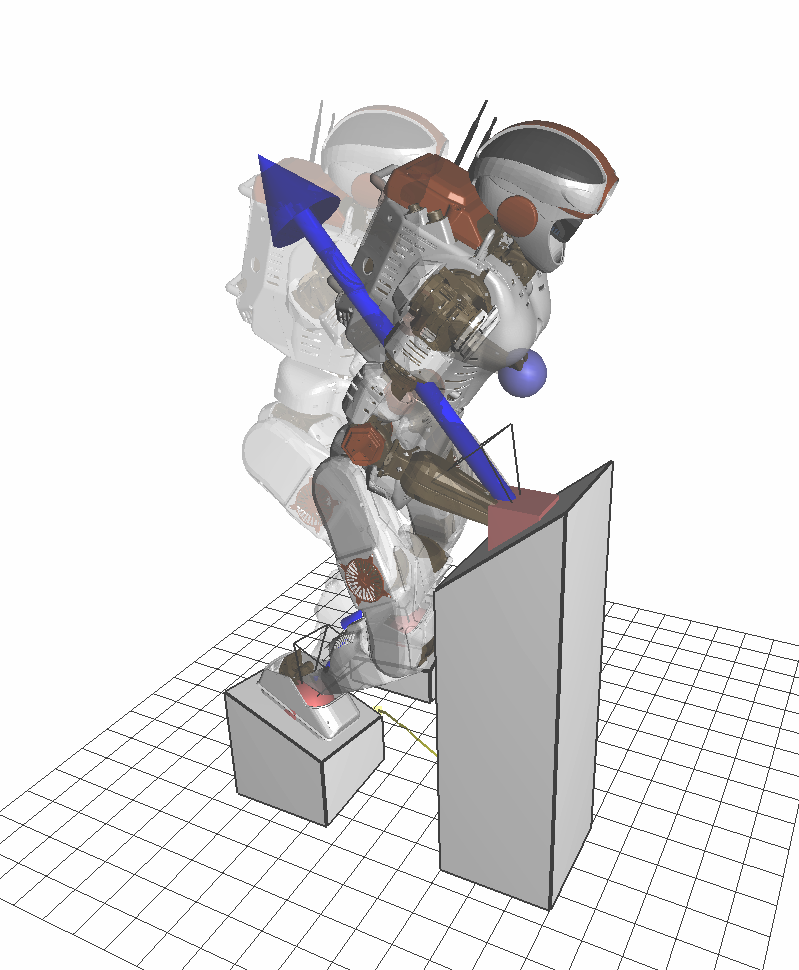}
		    \caption{Pushing}
	    \end{subfigure}
	    \begin{subfigure}{1.0\textwidth}
		    \centering
		    \includegraphics[height=2.2cm,trim={0 100 0 100},clip]{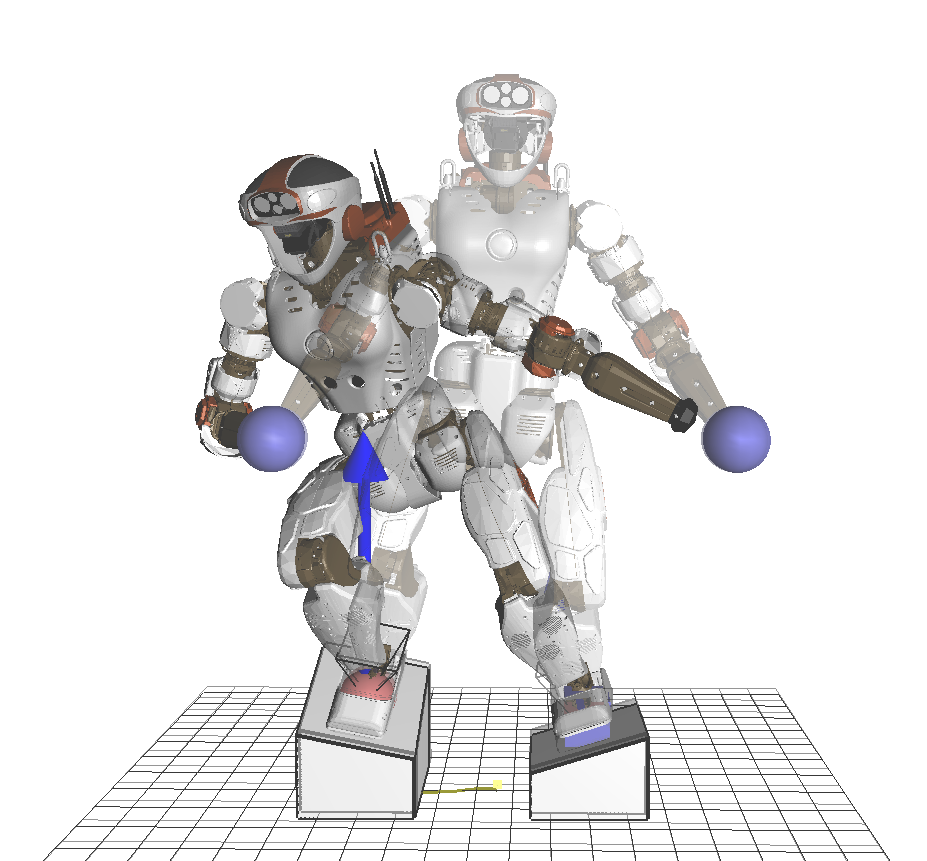}
		    \caption{Contact switching}
	    \end{subfigure}
    \end{minipage}
    \begin{minipage}{.83\textwidth}
    	\centering
	    \includegraphics[width=1.0\linewidth]{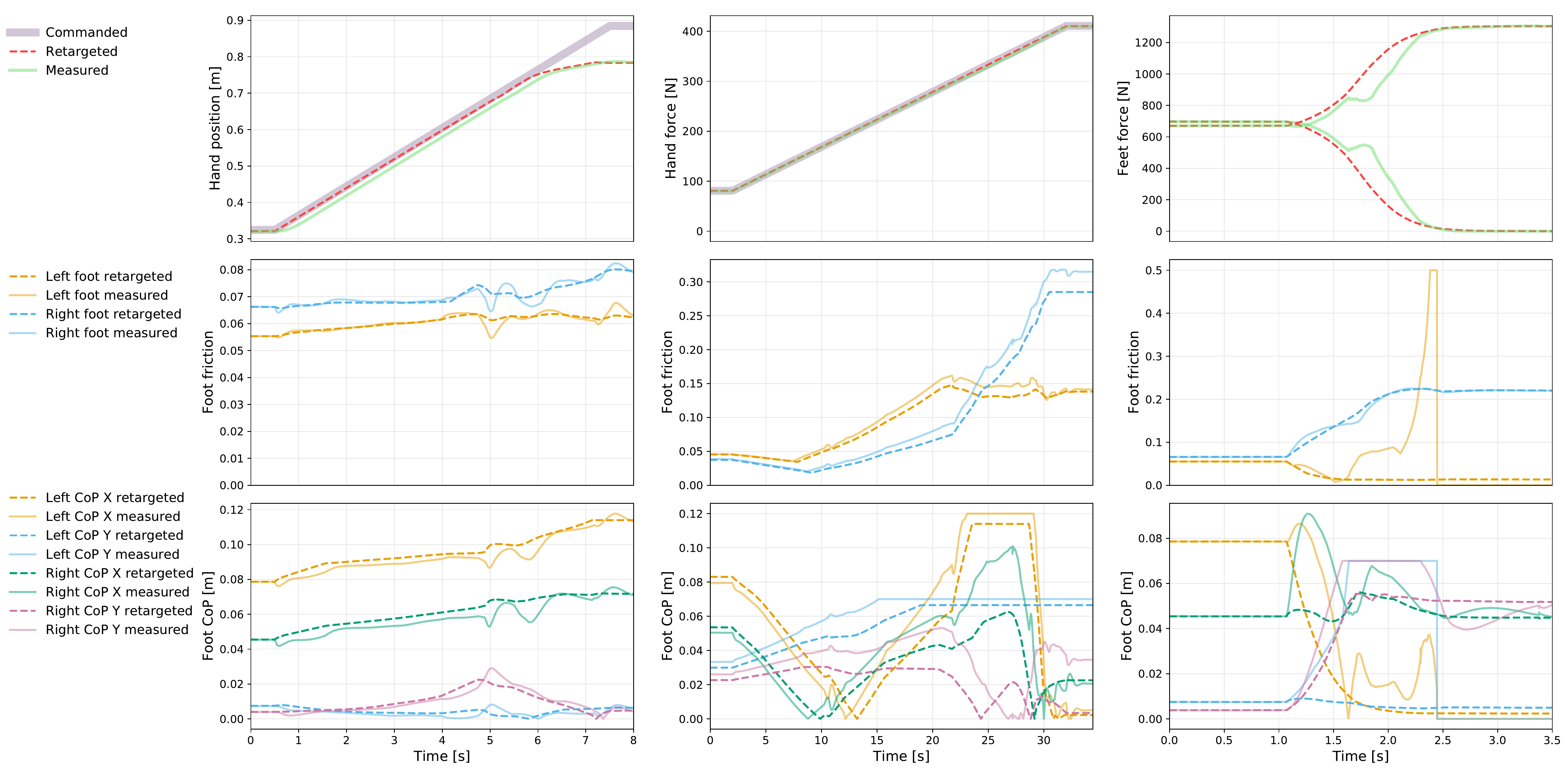}
	    \begin{subfigure}{0.4\linewidth}\end{subfigure}
	    \begin{subfigure}{0.35\linewidth}\caption*{\hspace{1cm}(d) Far reaching}\end{subfigure}
	    \begin{subfigure}{0.30\linewidth}\caption*{(e) Pushing}\end{subfigure}
	    \begin{subfigure}{0.23\linewidth}\caption*{\hspace{1cm}(f) Contact switching}\end{subfigure}
    \end{minipage}
	\caption{
        Multi-contact teleoperation of the Valkyrie robot in Pybullet simulation: (a) far reaching with the left hand; (b) pushing with the right hand; (c) contact switching to disengage and lift the left foot. We compare the operator's command, the desired configuration optimized by SEIKO and the measured one tracked by the dynamic controller, as shown in data plots from (d-f): (d) the position of the left hand during reaching; (e) the contact force of the right hand during pushing; (f) the force distribution among the two feet during the contact switching. For each task, the top row shows the retargeted and measured signals specific to the task, the middle row shows the friction ratio of the feet, and the bottom row shows the CoP.
    }
    \label{fig:valkyrie_figure}
\end{figure*}

\begin{figure*}[h!]
    \begin{minipage}{.2\textwidth}
        \begin{subfigure}{1.0\textwidth}
		    \centering
		    \includegraphics[height=2.2cm,trim={0 0 0 0},clip]{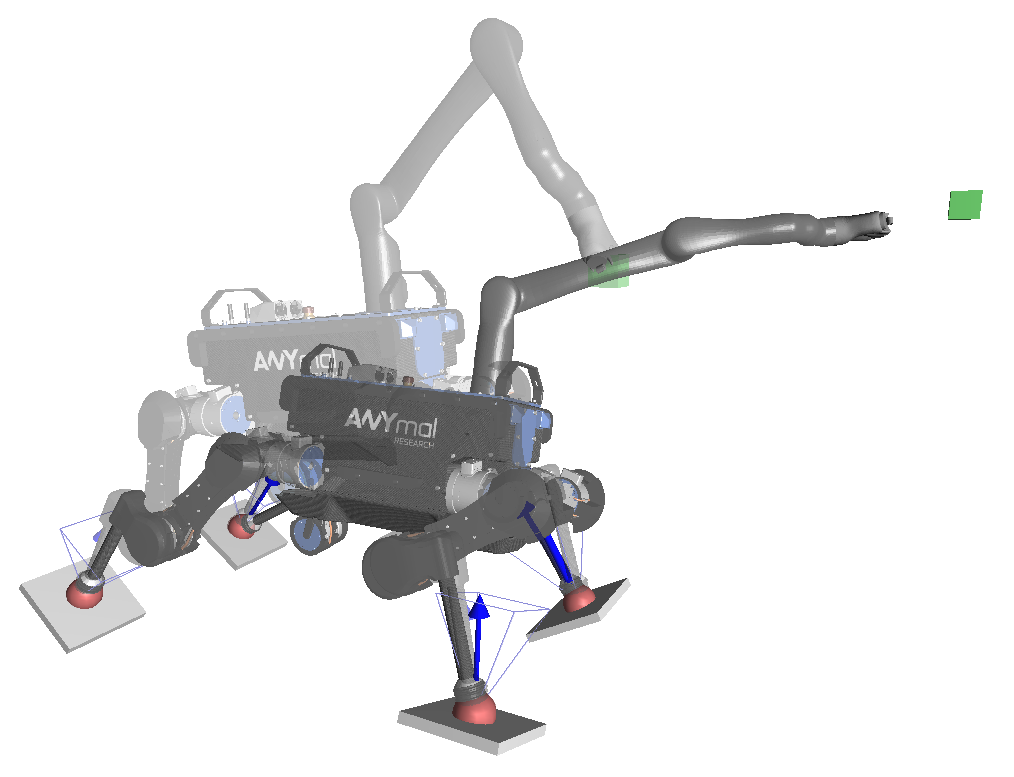}
		    \caption{Far reaching}
	    \end{subfigure}
	    \begin{subfigure}{1.0\textwidth}
		    \centering
		    \includegraphics[height=2.2cm,trim={0 0 0 0},clip]{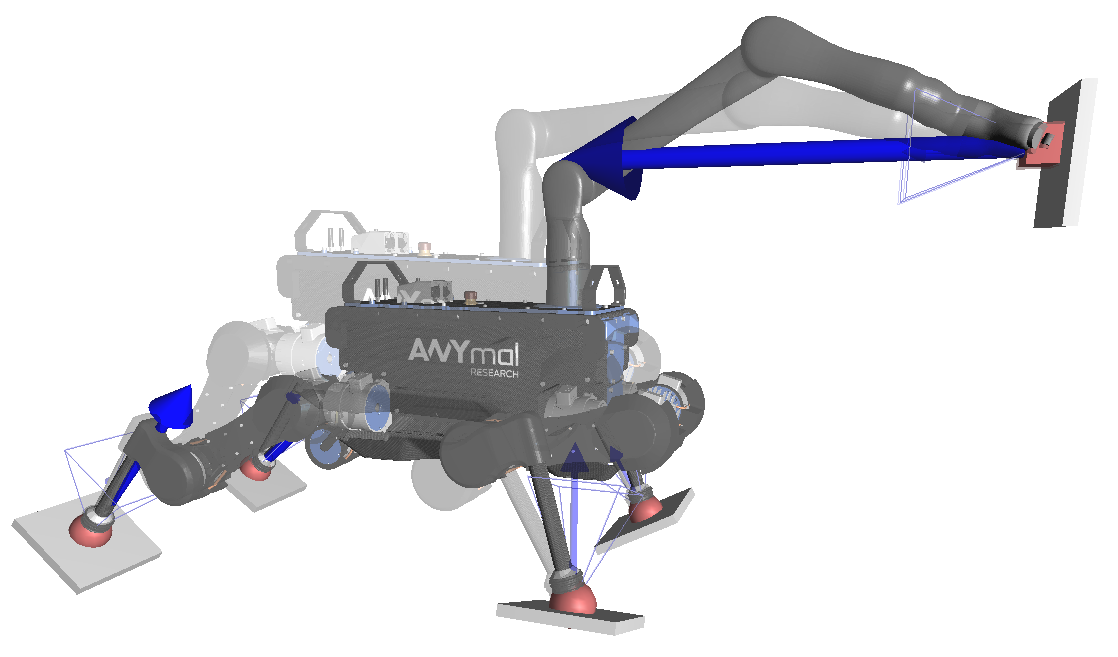}
		    \caption{Pushing}
	    \end{subfigure}
	    \begin{subfigure}{1.0\textwidth}
		    \centering
		    \includegraphics[height=2.2cm,trim={0 0 0 0},clip]{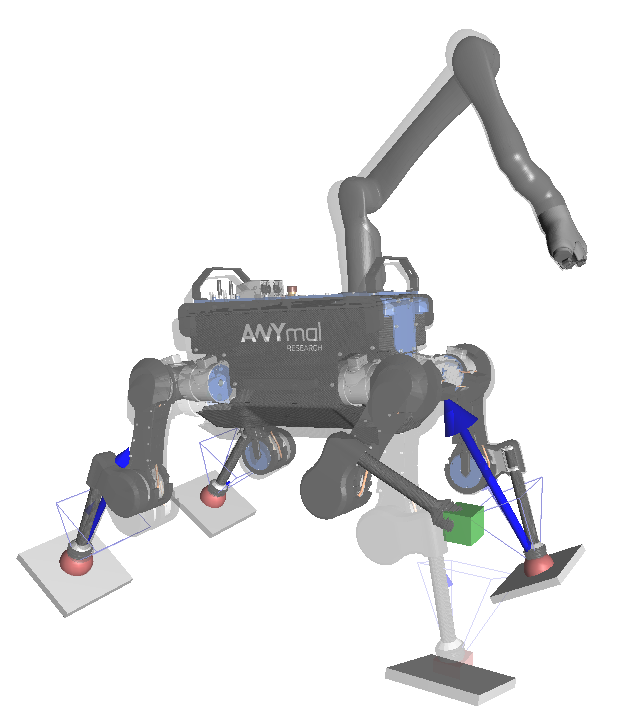}
		    \caption{Contact switching}
	    \end{subfigure}
    \end{minipage}
    \begin{minipage}{.79\textwidth}
    	\centering
	    \includegraphics[width=1.0\linewidth]{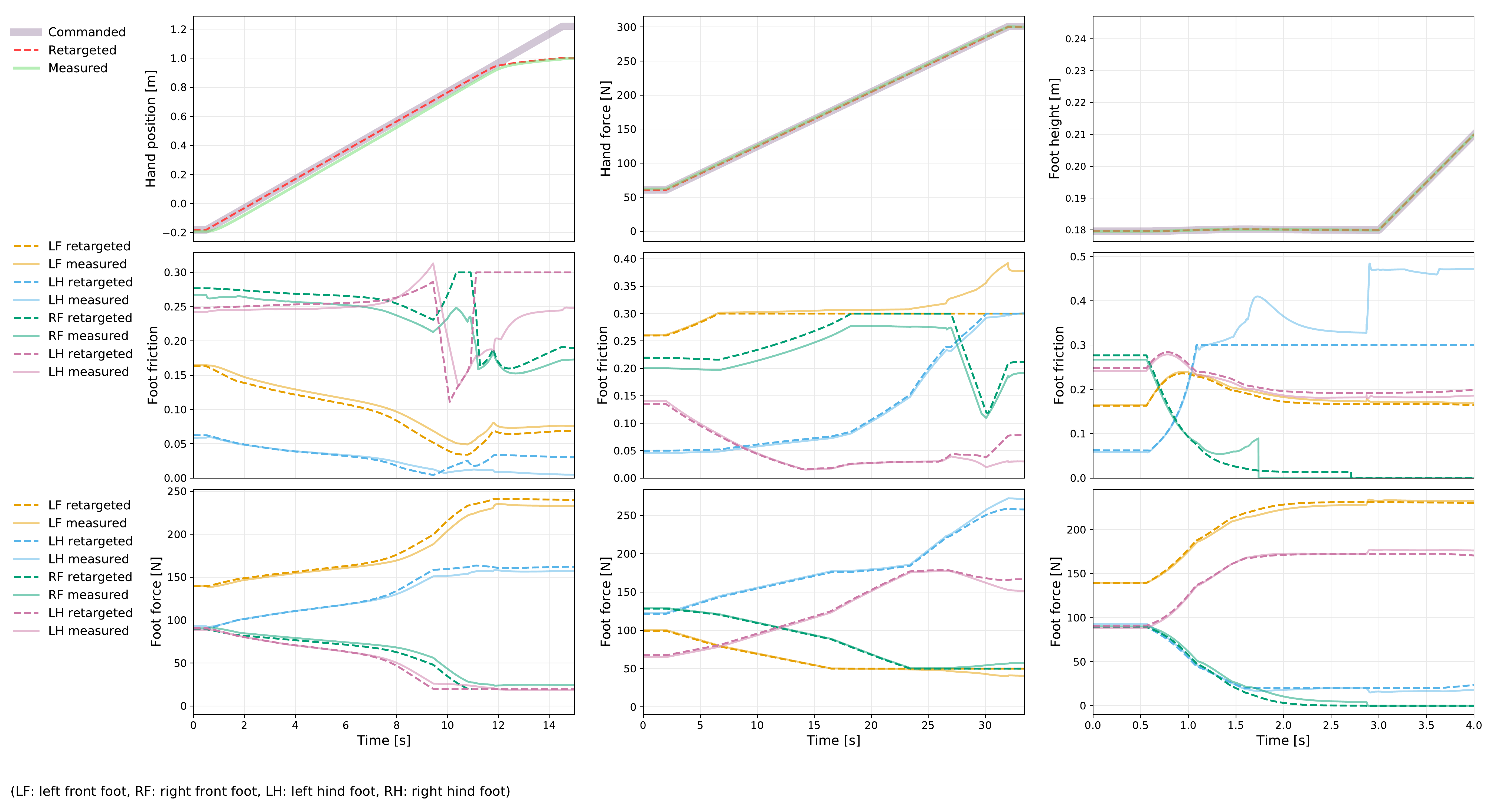}
	    \begin{subfigure}{0.4\linewidth}\end{subfigure}
	    \begin{subfigure}{0.35\linewidth}\caption*{\hspace{1cm}(d) Far reaching}\end{subfigure}
	    \begin{subfigure}{0.30\linewidth}\caption*{(e) Pushing}\end{subfigure}
	    \begin{subfigure}{0.23\linewidth}\caption*{\hspace{1cm}(f) Contact switching}\end{subfigure}
    \end{minipage}
	\caption{
	   Multi-contact teleoperation of the ANYmal robot in Pybullet simulation: (a) far reaching; (b) pushing; (c) contact switching to lift the front right foot (RF) at $\SI{3}{\second}$. We compare the operator's command, the desired configuration optimized by SEIKO and the measured one tracked by the dynamic controller, as shown in the data plots from (d-f): (d) the position of the hand during reaching; (e) the contact force of the hand during pushing; (f) the height of the front right foot during contact switching and lifting. For each task, the top row shows the retargeted and measured signals specific to the task, the middle row shows the friction ratio, and the bottom row shows the normal contact force distribution among the feet.
    }
    \label{fig:anymal_figure}
\end{figure*}

\begin{figure}[t]
    \centering
    \includegraphics[width=0.85\linewidth]{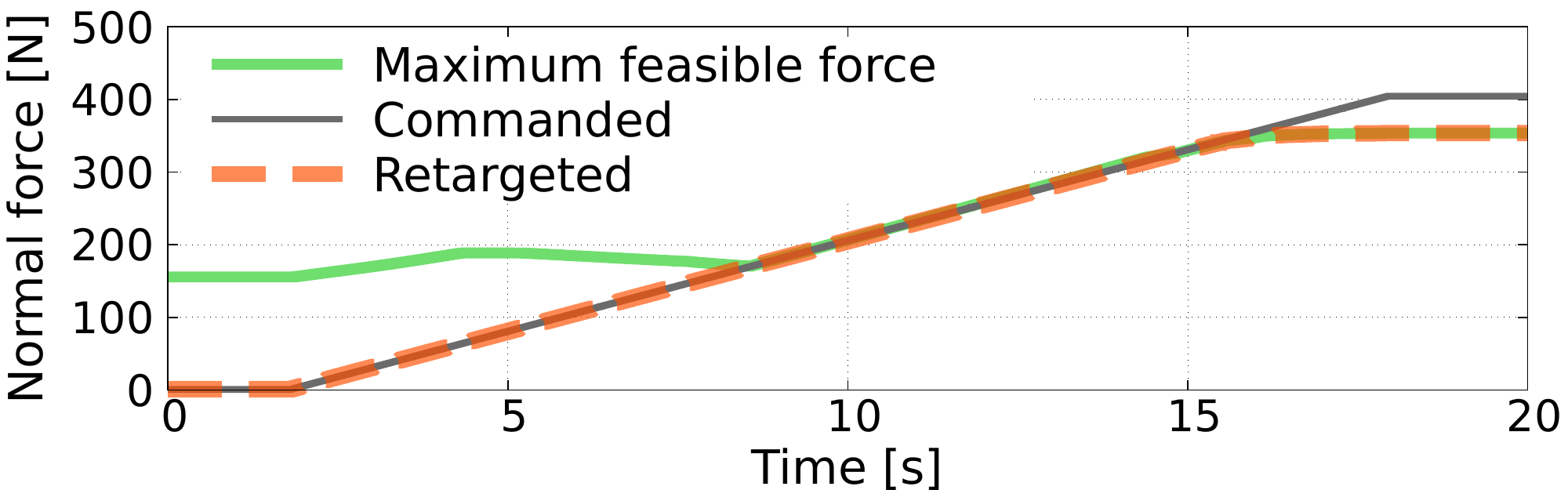}
    \caption{
        Contact force from Valkyrie's pushing task. Commanded and retargeted normal forces are compared with the maximum feasible force at the edge of the manipulability polytope.
    } \label{fig:plot_push}
\end{figure}

\begin{figure}[t]
    \centering
    \includegraphics[height=3.0cm]{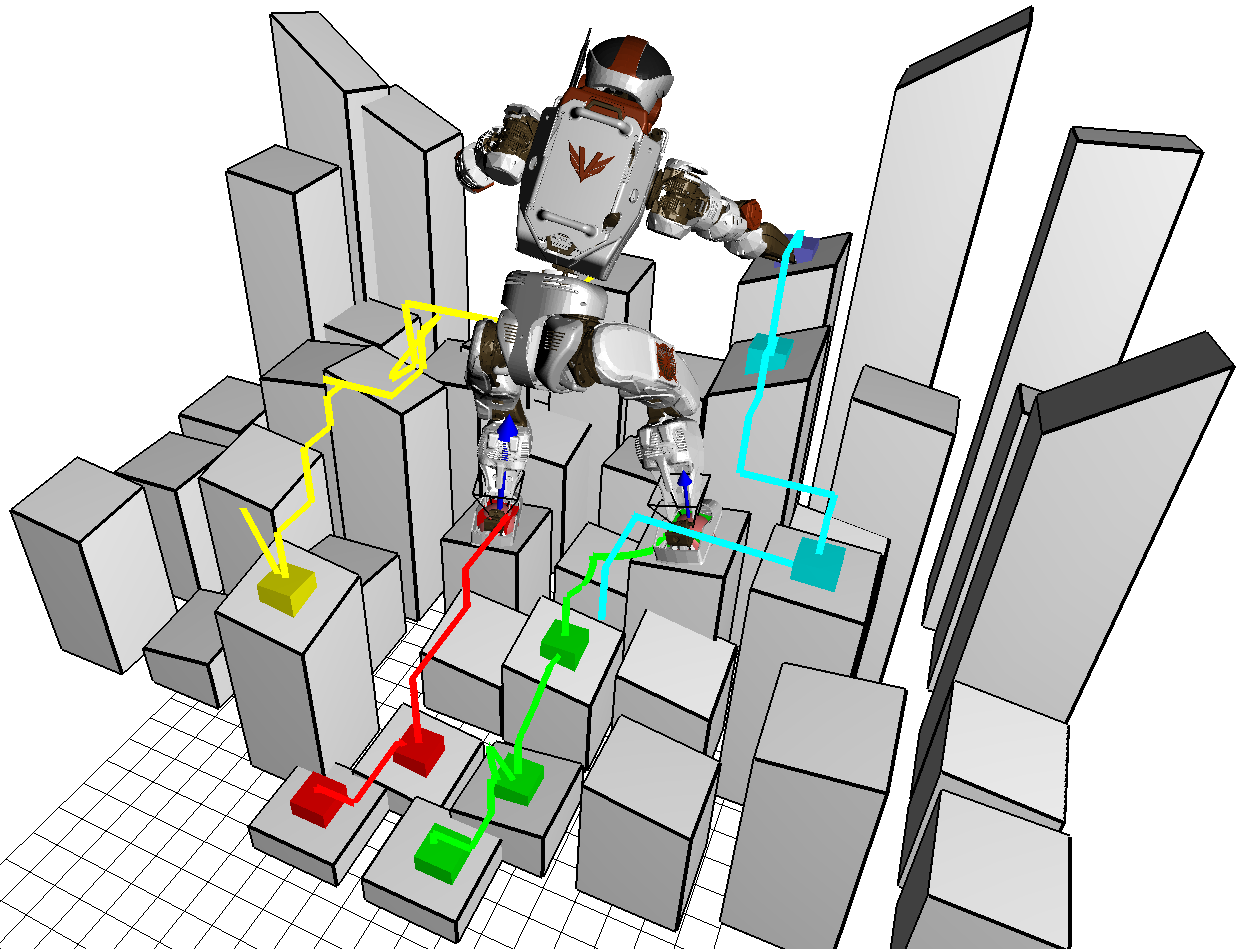}
    \includegraphics[height=3.0cm]{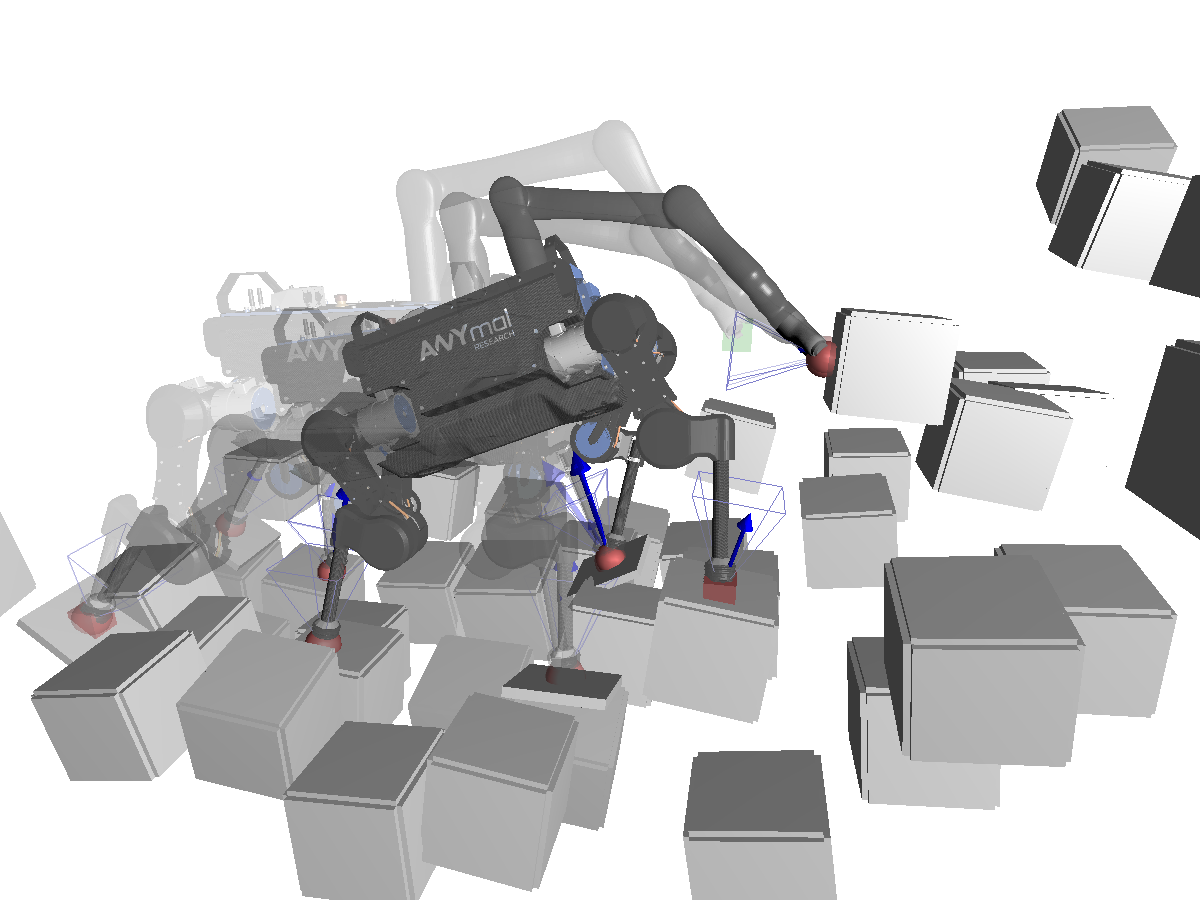}
    \caption{
        Teleoperated multi-contact locomotion on uneven terrains using hands and feet for Valkyrie and ANYmal robots. The operator can manually chose the contact sequence, command the motion of the end-effector and activate the contact transition.
    } \label{fig:parkour}
\end{figure}

The capabilities of our formulation are evaluated on four tasks:
(a) extreme reaching motions beyond the feasibility boundary, (b) hand pushing, (c) contact switching, and locomotion on complex uneven terrain.
We validated the whole control architecture in the Pybullet simulator, including SEIKO and a dynamic controller for tracking of the desired configuration.
Both SEIKO retargeting and the whole body dynamic controller run at $\SI{1000}{\hertz}$.
We show the teleoperation of Valkyrie in Fig.~\ref{fig:valkyrie_figure} and ANYmal in Fig.~\ref{fig:anymal_figure}.
For these experiments, the operator commanded the robot in real-time though a visualization and keyboard interface.

We evaluate the stability of the contacts using the Center of Pressure (CoP) expressed as $CoP_x = \frac{|\tau_y|}{f_z}, CoP_y = \frac{|\tau_x|}{f_z}$, and the friction ratio as $\eta = \frac{\mathsf{max}(|f_x|, |f_y|)}{f_z}$, where $\tau_x, \tau_y \in \R$ are the plane contact lateral torques, $f_x, f_x \in \R$ are the tangential contact forces and $f_z \in \R$ is the normal contact force. The non-tilting condition is satisfied for the feet of the humanoid, when $CoP_x < l_x, CoP_x < l_y$ and the non-sliding conditions is met when $\eta < \mu$, where $l_x,l_y \in \R$ are the foot plane lengths and $\mu \in \R$ is the friction coefficient limit. For the humanoid $l_x = \SI{0.11}{\meter}$, $l_y = \SI{0.07}{\meter}$ and $\mu = 0.5$; and for the point-foot quadruped, a more conservative value of $\mu = 0.3$ is chosen.

\subsubsection{Multi-contact reaching motion under constraints}

When the operator commands an extreme forward reaching motion (Fig.~\ref{fig:valkyrie_figure}(a) and Fig.~\ref{fig:anymal_figure}(a)), SEIKO updates the desired posture which is tracked by the whole body inverse dynamic controller. The desired and measured positions follow the commanded one until the saturation of some constraints (Section~\ref{sec:eq_cst}) blocks its further motion to prevent any balance or physical limits violation.
The CoP $X$ position of Valkyrie's left foot saturates at the foot edge at $\SI{11}{\centi\meter}$ (Fig.~{\ref{fig:valkyrie_figure}}(d), bottom plot) while the reaching motion of ANYmal is constrained by the minimum contact forces (Fig.~{\ref{fig:anymal_figure}}(d), bottom plot).

At every time step, SEIKO provides a statically balanced and feasible whole body configuration. 
The robot can come to rest at any time because of a constraint saturation or a commanded stop and still be safe.
Supplementary materials in Section~4 present additional results on the smooth adapted motions produced by SEIKO in response to discontinuous commands. Fast dynamic motions that violate the quasi-static assumption are analyzed and compared in Sections 6 and 11 in the supplementary materials. An additional application for retargeting human motion capture into the humanoid's morphology is also presented in supplementary materials Section~5.

\subsubsection{Pushing tasks and force manipulability}\label{sec:push}

The operator teleoperated pushing tasks with the right hand in Fig.~\ref{fig:valkyrie_figure}(b) and with the arm's hand in Fig.~\ref{fig:anymal_figure}(b)
through the commanded normal contact force.
We used the optional mode that can command of the contact force (see Section~\ref{sec:interface}).
The commanded force $\bm{\lambda}_{\text{hand}}^{\text{target}}$ in \eqref{eq:cost} was gradually increased by the operator and the associated weight $\bm{w}_{\text{contact,~hand}}$ was set to $10^4$ to prioritize the pushing task.

Fig.~{\ref{fig:valkyrie_figure}}(e) and Fig.~{\ref{fig:anymal_figure}}(e) show on upper row the commanded, retargeted and realized contact force.
At first the system is able to increase the applied force by only redistributing the forces among all the contacts with marginal posture change. When the commanded force in Fig.~\ref{fig:plot_push} reaches the \textit{maximum feasible normal force} (see supplementary materials Section~3 and \cite{orsolino2018application, 9341498}), SEIKO tends to update the whole posture by ``sliding along the constraints''. This postural adaptation increases the maximum feasible force and allows the robot to apply more force until being blocked further by the saturation of the kinematic or contact constraints.

\subsubsection{Contact switching}

The operator triggers the contact switching mechanism detailed in Section~\ref{sec:method_contact_switching} to smoothly remove and lift the left foot in Fig.~\ref{fig:valkyrie_figure}(c) and the front right foot in Fig.~\ref{fig:anymal_figure}(c).
The wrench penalty weight $w_{\text{contact}}$ increases exponentially from $1$ to $10^5$ and drives both the posture change and the force redistribution among the remaining contacts (Fig.~\ref{fig:valkyrie_figure}(f) upper plot and Fig.~\ref{fig:anymal_figure}(f) lower plot).
With one contact point removed, the saturation of the constraints (friction, CoP, minimal normal force) tends to increase.

%\changeadd{Note that by the definition, the friction ratio $\eta$ and the CoP is no longer valid when the contact force $f_z$ decreases to zero. When this happens, the small tracking errors of the contact force look amplified which explains the fluctuations observed on the friction ration and CoP curves as the contact is removed.}

\subsubsection{Locomotion}
The command of individual end-effector motions combined with contact switching allows the robot to locomote over uneven terrains, assuming static equilibrium. The operator selects the sequence and commands the location of the multi-contact stances by reaching, probing and triggering contact switch of the end-effectors. The sequences of locomotion for both robots are in Fig.~\ref{fig:parkour}, and the details of the retargeted constraints and tracking for the ANYmal robot can be found in Section~8 of the supplementary materials..

\section{Discussion}\label{sec:discussion}

The method we proposed is particularly suitable for interactive teleoperation where high level commands from the operator are retargeted for the robot's morphology. Unlike offline planners which computes the feasible trajectories offline, SEIKO computes feasible references online to be tracked by a dynamic controller. In slow motion cases, both SEIKO and existing approaches work equally well when commanded motions remain within the physical limits of the robot, because no safety precautions are actually required (see Section~10 of supplementary materials). However, when the limits of robots are violated, or the target position commands are infeasible due to operator's mistake, without SEIKO, the whole-body QP controller itself cannot maintain the robot's balance. In contrast, the proposed SEIKO can successfully restrict references within safety boundaries and guarantees the robot's long-term stability. 

SEIKO is applicable to multi-contact teleoperation cases which have not been solved by previous approaches, based on two main assumptions: the quasi-static equilibrium and the possible state discrepancy at contact initiation. The quasi-static assumption is required to guarantee the long-term balance because future commands are unknown in the context of teleoperation. Despite this assumption, we have demonstrated that our scheme still achieved acceptable operating velocities in conjunction with the whole-body QP controller (see Fig.~{\ref{fig:valkyrie_figure}}-{\ref{fig:anymal_figure}}). In Sections 6 and 11 of the supplementary material, we showed that the Valkyrie humanoid's hand can safely reach the velocity of $\SI{30}{cm/s}$ in reaching tasks, using parameters that trade off the maximum reachable distance for more conservative postures. Hence, SEIKO still works well for motions of moderate speeds, which suits for a large range of practical loco-manipulation applications, where feasibility and safety are more important than the speed.

The success of teleoperation tasks intrinsically puts dependencies on human motor skills, e.g., our work relies on humans for perception of the environment. To avoid the discrepancies between the model and the actual contact state that could destabilize the robot, the second assumption is that the human operator should command the robot's end-effector to be in contact with a new surface before triggering the contact switching. If the operator makes mistakes such as colliding the end-effector with the environment, the system will rely on the dynamic controller to attenuate such disturbances since the impact force is not part of SEIKO's formulation. Detecting and dealing with unexpected external force perturbations or collisions in the motion retargeting is a promising direction for future work.

SEIKO guarantees the generation of safe and feasible postures, and enhances the safety of intuitive, interactive multi-contact teleoperation, which is robust to human mistakes. This is an advantageous feature in teleoperation, especially when the communication latency can be a major source of human-factor risks. If operator's commands are erroneous due to an impeded communication link, the proposed SEIKO will automatically adapt and convert wrong commanded motions into viable solutions to ensure safety.

Extensive experiments in this work show that few adjustments are needed to transfer the parameters to a different robot once they are tuned (listed in \mbox{\autoref{table:ikid_parameters}}). These parameters have physical significance on how to affect the optimization outcome, so the adjustments are straightforward. Alternatively, Bayesian Optimization could be used for automatic tuning \mbox{\cite{yuan2019bayesian}}.

Similar to general nonlinear optimizations, the proposed scheme exhibits local minimum problems, e.g., robotic arms may occasionally have trouble in returning to their initial poses in a near-singular posture. We have mitigated this problem by regularizing the joint position to a default position as a low-weighted task (with parameter $\bm{w}_{\text{posture}}$).

\section{Conclusion and Future Work}\label{sec:conclusion}
This paper presents an optimization-based motion retargeting method which is suitable for teleoperation of quasi-static multi-contact tasks, such as loco-manipulation -- a combination of locomotion and manipulation. We proposed the Sequential Equilibrium and Inverse Kinematics Optimization (SEIKO) to map and adapt the operator's commands into feasible retargeted configurations, as well as a smooth contact switching and transitions for multi-contact tasks. This method has been applied to teleoperate both the Valkyrie humanoid and the ANYmal quadruped robot. The online teleoperation was achieved and validated in the Pybullet and Gazebo simulators, which demonstrated the effectiveness of the proposed method to guarantee the kinematic and dynamic feasibility.

As the future work, it would be beneficial to estimate external contact forces, which can be included as a bias vector in the equilibrium equation of SEIKO. Hence, we can automatically trigger contact switching when an end-effector pushes on the environment. Also, the presented formulation enforces all contact constraints and joint limits, without considering collisions. Since our retargeting formulation is compatible with self-collision avoidance as implemented in \cite{hoffman2018robot,di2016multi,vaillant2016multi}, collision avoidance can be a future extension as well.

\section*{Acknowledgment}

Authors would like to thank Justin Carpentier and the developers of the Pinocchio library, and Steve Tonneau for the valuable discussions. This research is supported by the EPSRC Future AI and Robotics for Space (EP/R026092/1), and EU Horizon2020 project Harmony (101017008).

%%%
%%% Bibliography
%%%

\bibliographystyle{templates/IEEEtran}
\bibliography{TeleOpIKID_quentin}

\begin{thebibliography}{10}
\providecommand{\url}[1]{#1}
\csname url@samestyle\endcsname
\providecommand{\newblock}{\relax}
\providecommand{\bibinfo}[2]{#2}
\providecommand{\BIBentrySTDinterwordspacing}{\spaceskip=0pt\relax}
\providecommand{\BIBentryALTinterwordstretchfactor}{4}
\providecommand{\BIBentryALTinterwordspacing}{\spaceskip=\fontdimen2\font plus
\BIBentryALTinterwordstretchfactor\fontdimen3\font minus
  \fontdimen4\font\relax}
\providecommand{\BIBforeignlanguage}[2]{{%
\expandafter\ifx\csname l@#1\endcsname\relax
\typeout{** WARNING: IEEEtran.bst: No hyphenation pattern has been}%
\typeout{** loaded for the language `#1'. Using the pattern for}%
\typeout{** the default language instead.}%
\else
\language=\csname l@#1\endcsname
\fi
#2}}
\providecommand{\BIBdecl}{\relax}
\BIBdecl

\bibitem{dedonato2015human}
M.~DeDonato, V.~Dimitrov, R.~Du, R.~Giovacchini, K.~Knoedler, X.~Long,
  F.~Polido, M.~A. Gennert, T.~Pad{\i}r, S.~Feng \emph{et~al.},
  ``Human-in-the-loop control of a humanoid robot for disaster response: A
  report from the darpa robotics challenge trials,'' \emph{Journal of Field
  Robotics}, vol.~32, no.~2, pp. 275--292, 2015.

\bibitem{johnson2015team}
M.~Johnson, B.~Shrewsbury, S.~Bertrand, T.~Wu, D.~Duran, M.~Floyd, P.~Abeles,
  D.~Stephen, N.~Mertins, A.~Lesman \emph{et~al.}, ``Team ihmc's lessons
  learned from the darpa robotics challenge trials,'' \emph{Journal of Field
  Robotics}, vol.~32, no.~2, pp. 192--208, 2015.

\bibitem{schmaus2018preliminary}
P.~Schmaus, D.~Leidner, T.~Kr{\"u}ger, A.~Schiele, B.~Pleintinger, R.~Bayer,
  and N.~Y. Lii, ``Preliminary insights from the meteron supvis justin
  space-robotics experiment,'' \emph{IEEE Robotics and Automation Letters},
  vol.~3, no.~4, pp. 3836--3843, 2018.

\bibitem{de2011set}
F.~B. de~Frescheville, S.~Martin, N.~Policella, D.~Patterson, M.~Aiple, and
  P.~Steele, ``Set-up and validation of meteron end-to-end network for robotic
  experiments,'' in \emph{ASTRA Conference}, 2011.

\bibitem{9145664}
I.~Chatzinikolaidis, Y.~You, and Z.~Li, ``Contact-implicit trajectory
  optimization using an analytically solvable contact model for locomotion on
  variable ground,'' \emph{IEEE Robotics and Automation Letters}, vol.~5,
  no.~4, pp. 6357--6364, 2020.

\bibitem{MontecilloPuente2010OnRW}
F.-J. Montecillo-Puente, M.~N. Sreenivasa, and J.-P. Laumond, ``On real-time
  whole-body human to humanoid motion transfer,'' in \emph{ICINCO}, 2010.

\bibitem{koenemann2014real}
J.~Koenemann, F.~Burget, and M.~Bennewitz, ``Real-time imitation of human
  whole-body motions by humanoids,'' in \emph{2014 IEEE International
  Conference on Robotics and Automation (ICRA)}.\hskip 1em plus 0.5em minus
  0.4em\relax IEEE, 2014, pp. 2806--2812.

\bibitem{darvish2019whole}
K.~Darvish, Y.~Tirupachuri, G.~Romualdi, L.~Rapetti, D.~Ferigo, F.~J.~A.
  Chavez, and D.~Pucci, ``Whole-body geometric retargeting for humanoid
  robots,'' in \emph{2019 IEEE-RAS 19th International Conference on Humanoid
  Robots (Humanoids)}.\hskip 1em plus 0.5em minus 0.4em\relax IEEE, 2019, pp.
  679--686.

\bibitem{abi2018humanoid}
F.~Abi-Farraj, B.~Henze, A.~Werner, M.~Panzirsch, C.~Ott, and M.~Roa,
  ``Humanoid teleoperation using task-relevant haptic feedback,'' in
  \emph{IEEE/RSJ Int. Conf. on Intelligent Robots and Systems, IROS'18}, 2018.

\bibitem{9117048}
J.~Oh, O.~Sim, B.~Cho, K.~Lee, and J.-H. Oh, ``Online delayed reference
  generation for a humanoid imitating human walking motion,'' \emph{IEEE/ASME
  Transactions on Mechatronics}, vol.~26, no.~1, pp. 102--112, 2021.

\bibitem{hoffman2018robot}
E.~M. Hoffman, A.~Rocchi, N.~G. Tsagarakis, and D.~G. Caldwell, ``Robot
  dynamics constraint for inverse kinematics,'' \emph{Advances in Robot
  Kinematics 2016}, vol.~4, p. 275, 2018.

\bibitem{vukobratovic2004zero}
M.~Vukobratovi{\'c} and B.~Borovac, ``Zero-moment point—thirty five years of
  its life,'' \emph{International journal of humanoid robotics}, vol.~1,
  no.~01, pp. 157--173, 2004.

\bibitem{kajita20013d}
S.~Kajita, F.~Kanehiro, K.~Kaneko, K.~Yokoi, and H.~Hirukawa, ``The 3d linear
  inverted pendulum mode: A simple modeling for a biped walking pattern
  generation,'' in \emph{Proceedings 2001 IEEE/RSJ International Conference on
  Intelligent Robots and Systems}, vol.~1.\hskip 1em plus 0.5em minus
  0.4em\relax IEEE, 2001, pp. 239--246.

\bibitem{yamane2003dynamics}
K.~Yamane and Y.~Nakamura, ``Dynamics filter-concept and implementation of
  online motion generator for human figures,'' \emph{IEEE transactions on
  robotics and automation}, vol.~19, no.~3, pp. 421--432, 2003.

\bibitem{ishiguro2017bipedal}
Y.~Ishiguro, K.~Kojima, F.~Sugai, S.~Nozawa, Y.~Kakiuchi, K.~Okada, and
  M.~Inaba, ``Bipedal oriented whole body master-slave system for dynamic
  secured locomotion with {LIP} safety constraints,'' in \emph{2017 IEEE/RSJ
  International Conference on Intelligent Robots and Systems (IROS)}, 2017, pp.
  376--382.

\bibitem{Penco2019Teleop}
L.~{Penco}, N.~{Scianca}, V.~{Modugno}, L.~{Lanari}, G.~{Oriolo}, and
  S.~{Ivaldi}, ``A multimode teleoperation framework for humanoid
  loco-manipulation: An application for the icub robot,'' \emph{IEEE Robotics
  Automation Magazine}, vol.~26, no.~4, pp. 73--82, Dec 2019.

\bibitem{del2018zero}
A.~Del~Prete, S.~Tonneau, and N.~Mansard, ``Zero step capturability for legged
  robots in multicontact,'' \emph{IEEE Transactions on Robotics}, vol.~34,
  no.~4, pp. 1021--1034, 2018.

\bibitem{7139908}
Z.~Li, C.~Zhou, J.~Castano, X.~Wang, F.~Negrello, N.~G. Tsagarakis, and D.~G.
  Caldwell, ``Fall prediction of legged robots based on energy state and its
  implication of balance augmentation: A study on the humanoid,'' in \emph{2015
  IEEE International Conference on Robotics and Automation (ICRA)}, 2015, pp.
  5094--5100.

\bibitem{di2016multi}
A.~Di~Fava, K.~Bouyarmane, K.~Chappellet, E.~Ruffaldi, and A.~Kheddar,
  ``Multi-contact motion retargeting from human to humanoid robot,'' in
  \emph{IEEE-RAS 16th International Conference on Humanoid Robots}, 2016, pp.
  1081--1086.

\bibitem{bretl2008testing}
T.~Bretl and S.~Lall, ``Testing static equilibrium for legged robots,''
  \emph{IEEE Transactions on Robotics}, vol.~24, no.~4, pp. 794--807, 2008.

\bibitem{shigematsu2019generating}
R.~Shigematsu, M.~Murooka, Y.~Kakiuchi, K.~Okada, and M.~Inaba, ``Generating a
  key pose sequence based on kinematics and statics optimization for
  manipulating a heavy object by a humanoid robot,'' in \emph{2019 IEEE/RSJ
  International Conference on Intelligent Robots and Systems (IROS)}.\hskip 1em
  plus 0.5em minus 0.4em\relax IEEE, 2019, pp. 3852--3859.

\bibitem{bouyarmane2012humanoid}
K.~Bouyarmane and A.~Kheddar, ``Humanoid robot locomotion and manipulation step
  planning,'' \emph{Advanced Robotics}, vol.~26, no.~10, pp. 1099--1126, 2012.

\bibitem{brossette2018multicontact}
S.~Brossette, A.~Escande, and A.~Kheddar, ``Multicontact postures computation
  on manifolds,'' \emph{IEEE Transactions on Robotics}, 2018.

\bibitem{feng2014optimization}
S.~Feng, E.~Whitman, X.~Xinjilefu, and C.~G. Atkeson, ``Optimization based full
  body control for the atlas robot,'' in \emph{2014 IEEE-RAS International
  Conference on Humanoid Robots}, 2014.

\bibitem{autonrobot2016}
Z.~Li, C.~Zhou, N.~Tsagarakis, and D.~Caldwell, ``Compliance control for
  stabilizing the humanoid on the changing slope based on terrain inclination
  estimation,'' \emph{Autonomous Robots}, vol.~40, 08 2016.

\bibitem{1525021}
E.~S. Neo, K.~Yokoi, S.~Kajita, F.~Kanehiro, and K.~Tanie, ``A switching
  command-based whole-body operation method for humanoid robots,''
  \emph{IEEE/ASME Transactions on Mechatronics}, vol.~10, no.~5, pp. 546--559,
  2005.

\bibitem{featherstone2014rigid}
R.~Featherstone, \emph{Rigid body dynamics algorithms}.\hskip 1em plus 0.5em
  minus 0.4em\relax Springer, 2014.

\bibitem{mansard2009versatile}
N.~Mansard, O.~Stasse, P.~Evrard, and A.~Kheddar, ``A versatile generalized
  inverted kinematics implementation for collaborative working humanoid robots:
  The stack of tasks,'' in \emph{2009 International Conference on Advanced
  Robotics}.\hskip 1em plus 0.5em minus 0.4em\relax IEEE, 2009, pp. 1--6.

\bibitem{rocchi2015opensot}
A.~Rocchi, E.~M. Hoffman, D.~G. Caldwell, and N.~G. Tsagarakis, ``Opensot: A
  whole-body control library for the compliant humanoid robot coman,'' in
  \emph{2015 IEEE International Conference on Robotics and Automation (ICRA)},
  2015.

\bibitem{gill2012sequential}
P.~E. Gill and E.~Wong, ``Sequential quadratic programming methods,'' in
  \emph{Mixed integer nonlinear programming}.\hskip 1em plus 0.5em minus
  0.4em\relax Springer, 2012, pp. 147--224.

\bibitem{kraft1988software}
D.~Kraft \emph{et~al.}, ``A software package for sequential quadratic
  programming,'' 1988.

\bibitem{caron2015stability}
S.~Caron, Q.-C. Pham, and Y.~Nakamura, ``Stability of surface contacts for
  humanoid robots: Closed-form formulae of the contact wrench cone for
  rectangular support areas,'' in \emph{2015 IEEE International Conference on
  Robotics and Automation (ICRA)}, 2015.

\bibitem{herzog2016momentum}
A.~Herzog, N.~Rotella, S.~Mason, F.~Grimminger, S.~Schaal, and L.~Righetti,
  ``Momentum control with hierarchical inverse dynamics on a torque-controlled
  humanoid,'' \emph{Autonomous Robots}, vol.~40, no.~3, pp. 473--491, 2016.

\bibitem{Felis2016}
M.~L. Felis, ``Rbdl: an efficient rigid-body dynamics library using recursive
  algorithms,'' \emph{Autonomous Robots}, pp. 1--17, 2016.

\bibitem{carpentier2019pinocchio}
J.~Carpentier, G.~Saurel, G.~Buondonno, J.~Mirabel, F.~Lamiraux, O.~Stasse, and
  N.~Mansard, ``The pinocchio c++ library -- a fast and flexible implementation
  of rigid body dynamics algorithms and their analytical derivatives,'' in
  \emph{IEEE International Symposium on System Integrations (SII)}, 2019.

\bibitem{carpentier2018analytical}
J.~Carpentier and N.~Mansard, ``Analytical derivatives of rigid body dynamics
  algorithms,'' in \emph{Robotics: Science and Systems (RSS 2018)}, 2018.

\bibitem{goldfarb1983numerically}
D.~Goldfarb and A.~Idnani, ``A numerically stable dual method for solving
  strictly convex quadratic programs,'' \emph{Mathematical programming},
  vol.~27, no.~1, pp. 1--33, 1983.

\bibitem{orsolino2018application}
R.~Orsolino, M.~Focchi, C.~Mastalli, H.~Dai, D.~G. Caldwell, and C.~Semini,
  ``Application of wrench-based feasibility analysis to the online trajectory
  optimization of legged robots,'' \emph{IEEE Robotics and Automation Letters},
  vol.~3, no.~4, pp. 3363--3370, 2018.

\bibitem{9341498}
W.~J. Wolfslag, C.~McGreavy, G.~Xin, C.~Tiseo, S.~Vijayakumar, and Z.~Li,
  ``Optimisation of body-ground contact for augmenting the whole-body
  loco-manipulation of quadruped robots,'' in \emph{2020 IEEE/RSJ International
  Conference on Intelligent Robots and Systems (IROS)}, 2020, pp. 3694--3701.

\bibitem{yuan2019bayesian}
K.~Yuan, I.~Chatzinikolaidis, and Z.~Li, ``Bayesian optimization for whole-body
  control of high-degree-of-freedom robots through reduction of
  dimensionality,'' \emph{IEEE Robotics and Automation Letters}, vol.~4, no.~3,
  pp. 2268--2275, 2019.

\bibitem{vaillant2016multi}
J.~Vaillant, A.~Kheddar, H.~Audren, F.~Keith, S.~Brossette, A.~Escande,
  K.~Bouyarmane, K.~Kaneko, M.~Morisawa, P.~Gergondet \emph{et~al.},
  ``Multi-contact vertical ladder climbing with an hrp-2 humanoid,''
  \emph{Autonomous Robots}, vol.~40, no.~3, pp. 561--580, 2016.

\end{thebibliography}

%\balance

%\vspace{-1.9cm}
\begin{IEEEbiographynophoto}{Quentin Rouxel}
is currently a post-doctoral researcher at the School of Informatics at the University of Edinburgh.
He received his PhD degree in 2017 at the University of Bordeaux in the robotic team Rhoban and the computer science laboratory of Bordeaux (LaBRI) and participated several years to the RoboCup competition in the humanoid kid-size league.
His research interests are in control and learning for legged and multi-contacts robotic systems.
\end{IEEEbiographynophoto}

\vskip -2\baselineskip plus -1fil
\begin{IEEEbiography}[{\includegraphics[width=1in,height=1.25in,clip,keepaspectratio]{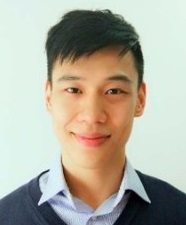}}]{Kai Yuan}
received his M.Sc. from the University of Edinburgh in Robotics and Autonomous Systems, and an M.Sc. and B.Sc. in Engineering Cybernetics from the University of Stuttgart. 

He is currently a PhD student at the University of Edinburgh with research interests in the control and optimisation of robots and the application of Machine Learning to further enhance the autonomy and intelligence of robots.
\end{IEEEbiography}

\vskip -2\baselineskip plus -1fil
\begin{IEEEbiography}[{\includegraphics[width=1in,height=1.25in,clip,keepaspectratio]{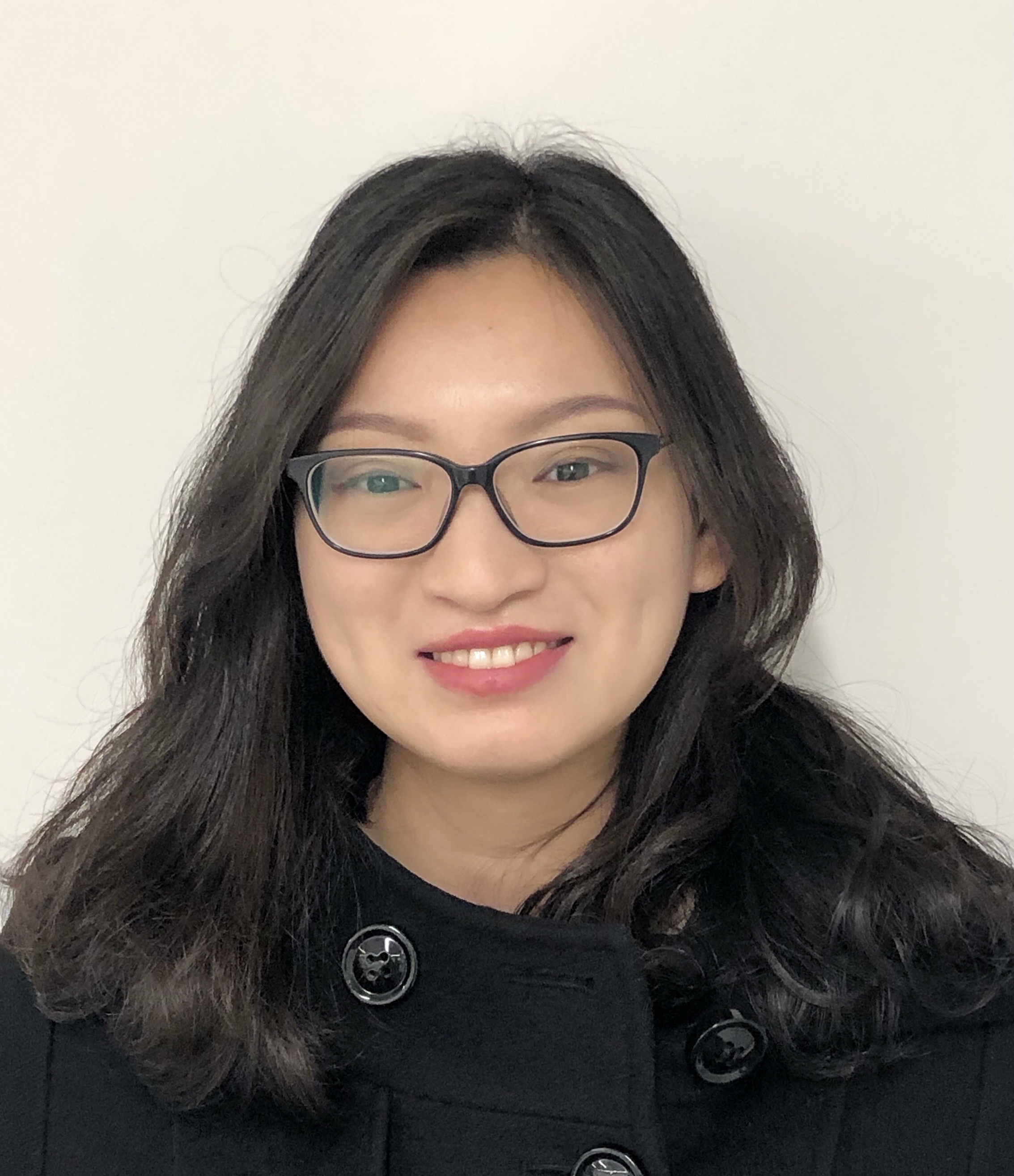}}]{Ruoshi Wen}is currently a postdoctoral research associate in the School of Informatics at the University of Edinburgh. Her research interests include human robot interaction and collaboration, biomedical signal processing, Bayesian inference, and machine learning. She is dedicated to enhance the grasping and manipulation ability of robots by integrating human intelligence.
\end{IEEEbiography}

\vskip -2\baselineskip plus -1fil
\begin{IEEEbiography}[{\includegraphics[width=1in,height=1.25in,clip,keepaspectratio]{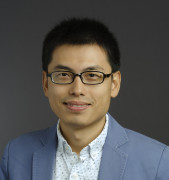}}]{Zhibin (Alex) Li} is an assistant professor at the School of Informatics, University of Edinburgh. He obtained an joint PhD degree in Robotics at the Italian Institute of Technology (IIT) and University of Genova in 2012. His research interests are in creating intelligent behaviors of dynamical systems with human comparable abilities to move and manipulate by inventing new control, optimization and deep learning technologies.
\end{IEEEbiography}

\end{document}